%% file: main.tex
\documentclass{article} 
\usepackage{collas2022_conference,times}

\input{math_commands.tex}

\usepackage{hyperref}
\hypersetup{
    colorlinks=true,
    linkcolor=red,
    filecolor=magenta,      
    urlcolor=blue,
    citecolor=purple,
    pdftitle={Overleaf Example},
    pdfpagemode=FullScreen,
    }
\usepackage{url}            
\usepackage{booktabs}       
\usepackage{amsfonts}       
\usepackage{nicefrac}       
\usepackage{microtype}      
\usepackage{xcolor}         
\usepackage{graphicx}
\usepackage{multirow} 
\usepackage[normalem]{ulem}
\usepackage{algorithm, algorithmic}
\usepackage{cancel}

\useunder{\uline}{\ul}{}
\usepackage{bookmark, units, RobStd, amsmath} 
\usepackage{caption, subcaption, units} 
\usepackage{enumitem, wrapfig}

\newcommand{\tablefont}[0]{\footnotesize}

\graphicspath{{images/}}

\title{How do Quadratic Regularizers Prevent Catastrophic Forgetting: The Role of Interpolation}
\input{title}

\collasfinalcopy 

\begin{document}

\maketitle

\input{files/abstract}
\input{files/intro}
\input{files/related}
\input{files/quadreg}

\input{files/limitations}
\input{files/explicitreg}

\input{files/experiments}
\input{files/conclusion}

\bibliography{main}
\bibliographystyle{collas2022_conference}

\input{appendix}

\end{document}

%% file: math_commands.tex
\usepackage{amsmath,amsfonts,bm}

\def\eqref#1{equation~\ref{#1}}

\def\1{\bm{1}}

\DeclareMathAlphabet{\mathsfit}{\encodingdefault}{\sfdefault}{m}{sl}
\SetMathAlphabet{\mathsfit}{bold}{\encodingdefault}{\sfdefault}{bx}{n}



%% file: title.tex
\author{%
  Ekdeep Singh~Lubana\\
  \texttt{eslubana@umich.edu} \\
  \And
  Puja Trivedi \\
  \texttt{pujat@umich.edu} \\
  \And
  Danai Koutra \\
  \texttt{dkoutra@umich.edu} \\
  \And
  Robert P.\ Dick \\
  \texttt{dickrp@umich.edu} \vspace{5pt}\\
  EECS Department\\
  University of Michigan\\
  }

\author{Ekdeep Singh Lubana, Puja Trivedi, Danai Koutra, \& Robert P. Dick \\
EECS Department\\
University of Michigan, Ann Arbor\\
USA \\
\texttt{\{eslubana, pujat, danaik, dickrp\}@umich.edu} \\
}

%% file: files/abstract.tex
\begin{abstract}
Catastrophic forgetting undermines the effectiveness of deep neural networks (DNNs) in scenarios such as continual learning and lifelong learning. While several methods have been proposed to tackle this problem, there is limited work explaining why these methods work well. This paper has the goal of better explaining a popularly used technique for avoiding catastrophic forgetting: quadratic regularization. We show that quadratic regularizers prevent forgetting of past tasks by interpolating current and previous values of model parameters at every training iteration. Over multiple training iterations, this interpolation operation reduces the learning rates of more important model parameters, thereby minimizing their movement. Our analysis also reveals two drawbacks of quadratic regularization: (a) dependence of parameter interpolation on training hyperparameters, which often leads to training instability and (b) assignment of lower importance to deeper layers, which are generally the place forgetting occurs in DNNs. Via a simple modification to the order of operations, we show these drawbacks can be easily avoided, resulting in 6.2\% higher average accuracy at 4.5\% lower average forgetting. We confirm the robustness of our results by training over 2000 models in different settings.
 \footnote{Code available at \url{https://github.com/EkdeepSLubana/QRforgetting}}
\end{abstract}

%% file: files/intro.tex
\section{Introduction}
\label{sec:intro}
Learning algorithms are often designed under the assumption of independent and identical data distributions. However, this assumption is violated in several practical scenarios, such as continual learning and lifelong learning, where data distributions evolve constantly. In such settings, deep neural networks (DNNs) witness catastrophic forgetting and have difficulty adapting to new tasks without losing performance on previously learned ones. Several past works have tried to address this problem. For example, \cite{EWC, MAS, SI, RWalk} propose quadratic regularizers that penalize changes in parameters which are important for preserving performance on past tasks; \cite{er_reservoir, mer, er_grad, dgr} describe the use of memory buffers or generative models to fine-tune on samples from past tasks while learning new ones; and \cite{PNN, cl_nas, den} propose dynamic modification of network architecture to increase model capacity for accommodating new tasks. While these works have shown promising results, a detailed understanding of their proposed methods is still to be developed. Understanding the reasons due to which existing methods for preventing catastrophic forgetting work or fail can open the possibility of developing better methods. 

With this motivation, in this work, we analyze \emph{quadratic regularization}, a popular technique for preventing catastrophic forgetting in DNNs. Specifically, quadratic regularization based methods penalize changes in model parameters that are important for maintaining performance on previously learned tasks. For example, if $\theta_{n}$ denotes model parameters corresponding to the $n^{\text{th}}$ task, $T_{n}$, then the total training loss under quadratic regularization follows
\begin{equation}
\label{eq:loss}
L = L_{T_{n}} + \frac{\lambda}{2} \sum_{k} \alpha_{n-1}^{(k)} \left(\theta_{n}^{(k)} - \theta_{n-1}^{*(k)}\right)^{2},
\end{equation}
where $L_{T_{n}}$ is the $n^{\text{th}}$ task's loss, $\lambda$ is a regularization constant, $\theta_{n-1}^{*}$ is model parameterization at the end of $(n-1)^{\text{th}}$ task, $\alpha_{n-1}$ is the importance of parameters according to tasks $0$ to $n-1$, and $v^{(k)}$ indexes a vector $v$ at location $k$. Different methods define importance differently. For ex., \cite{MAS} define importance as the sensitivity of model output to changes in its parameters, while \cite{EWC} define importance as the diagonal of Fisher information matrix.

Beyond continual/lifelong learning and related applications, quadratic regularizers are often used for improving performance in transfer learning scenarios (\cite{finetune}). However, despite their clear importance, there is limited work explaining why quadratic regularizers reduce catastrophic forgetting (see \cite{raghu, taylor_reg} for notable exceptions). To bridge this gap, we analyze parameter updates under quadratic regularization and show it prevents forgetting by \emph{interpolating current values of model parameters and their values at the end of the previous task's training} (see \autoref{eq:param_update}). By unrolling this interpolation operation over multiple iterations, we further show that, \emph{in proportion to their importance, quadratic regularizers change the ``effective'' learning rate of a parameter} (\autoref{eq:unrolled}): learning rate of parameters that are more important to previously learned tasks is reduced, disallowing their change; meanwhile parameters that are less important are allowed to change relatively freely, allowing the model to accommodate new tasks.

Our analysis exposes two important drawbacks to using quadratic regularization for preventing catastrophic forgetting. (i) We find improper hyperparameter configurations can result in \emph{extrapolation}, instead of interpolation, of parameters (see \autoref{sec:quadreg}). This often leads to unstable training (see \autoref{sec:limitations}). (ii) To satisfy conservation properties associated with the hierarchical nature of DNNs (\cite{du, mechanics}), generally used importance definitions in quadratic regularization assign lower importance values to parameters in deeper layers (see \autoref{sec:limitations}). As recently shown by \cite{raghu}, forgetting in DNNs is primarily caused by changes to deeper layers' parameters. Thus, assignment of lower importance to deeper layers reduces the effectiveness of quadratic regularizers. Interestingly, both these limitations can be eliminated by breaking the update into two stages, thus making interpolation of parameters an explicit operation (see \autoref{sec:emr}).

\textbf{Main Contributions:} In this work, we analyze quadratic regularization based methods for mitigating catastrophic forgetting. We train more than 2000 models to verify our analysis. Our main contributions follow.
\begin{itemize}[leftmargin=*]
\item \textbf{Role of Interpolation in Quadratic Regularizers (\autoref{sec:quadreg}).} We analyze parameter updates under quadratic regularization and show that it interpolates current and past values of model parameters to prevent catastrophic forgetting at a given training iteration. Unrolled over multiple iterations, this interpolation operation reduces the effective learning rate of parameters that are more important for preserving performance on previously learned tasks. 
\item \textbf{Limitations in Quadratic Regularization (\autoref{sec:limitations}).} We identify two primary drawbacks in the formulation of quadratic regularization: (a) due to dependence of the interpolation operation on training hyperparameters, improper hyperparameter configurations often result in extrapolation of parameters (instead of interpolation), hence resulting in training instability and (b) assignment of lower importance to parameters in deeper layers, a consequence of the hierarchical nature of DNNs, makes it difficult for quadratic regularizers to prevent catastrophic forgetting. 
\item \textbf{Avoiding Limitations via Explicit Interpolation (\autoref{sec:emr}).} By making the interpolation operation explicit, we show training instability issues caused by sensitivity to training hyperparameters can be completely avoided. Further, we show that redefining importance scores relative to past tasks prevents catastrophic forgetting caused by inappropriate changes to deeper layer parameters. Overall, using these modifications enable quadratic regularizers to achieve 6.2\% higher average accuracy at 4.5\% lower average forgetting.
\end{itemize}


%% file: files/related.tex
\section{Related Work}
\label{sec:related}

Several prior works have proposed solutions to the problem of catastrophic forgetting in DNNs. Due to their higher relevance to our work, we focus on quadratic regularization based methods in this section. Detailed discussion of other techniques is provided in the appendix. 

\textbf{Quadratic Regularization Based Methods:} \cite{EWC} developed Elastic Weight Consolidation (EWC), a quadratic regularization strategy that defines the importance of model parameters using the diagonal of the Fisher information matrix. Since EWC, several quadratic regularizers have been proposed. For ex., \cite{SI} describe Synaptic Intelligence (SI), which defines a parameter's importance as the contribution of that parameter to reduction in loss for previous tasks; \cite{ritter} replace EWC's diagonal Fisher approximation with a block-diagonal Hessian approximation; \cite{MAS} describe Memory Aware Synapses (MAS), for which a parameter's importance is defined as the sensitivity of model output to change in that parameter; and \cite{RWalk} describe Riemannian Walk (RWalk), which defines a parameter's importance as the sum of the diagonal of the Fisher information matrix and the contribution of that parameter to reduction in loss for previous tasks. 

\textbf{Understanding Methods for Mitigating Catastrophic Forgetting:} A few papers have sought to understand the reasons behind catastrophic forgetting and the effectiveness of existing methods for alleviating it (\cite{genforget, regimes, lmc, ntk, nphard, ogd_guarantees}). We specifically highlight an important relevant work by \cite{raghu}, who use Centered Kernel Alignment (CKA) (\cite{CKA}) to measure representational similarity between models trained on different numbers of tasks. The authors demonstrate that catastrophic forgetting primarily arises due to adaptation of deeper layers to new tasks. Solutions that prevent catastrophic forgetting minimize this adaptation, thus ensuring the model can perform well on both novel and previously learned tasks. Closely related to our work is also the recent paper by \cite{taylor_reg}, who analyze quadratic regularizers by focusing on their asymptotic properties (i.e., convergence and generalization behavior). Similarly, \cite{benzing} discusses how different importance defintions in popular quadratic regularizers are related to each other via different approximations of the Fisher information matrix. In contrast to these papers, we focus on the model updates themselves: our work aims to understand limitations in quadratic regularization methods arising from its gradient descent dynamics. As we show, an implicit interpolation operation helps QR methods prevent catastrophic forgetting, and its interplay with training hyperparameters yields clear regimes of training instability in QR methods. We also note that some prior works have proposed Bayesian perspectives for understanding and improving quadratic regularization techniques by introducing better importance definitions (\cite{huszar, EWC, ritter}), however these works often make strong assumptions that are unlikely to hold in practice, such as assuming the model gradient remains zero for previous tasks (to allow a Laplace or second-order Taylor’s approximation). In contrast, our work makes no assumptions about the model’s capabilities for solving a prior task, using the exact dynamics equations. 


%% file: files/quadreg.tex
\section{The Role of Interpolation in Quadratic Regularization}
\label{sec:quadreg}
We first establish notations used throughout the paper. $\theta_{n}$ denotes model parameters and $L_{T_{n}}$ denotes a task-specific loss for the $n^{\text{th}}$ task; $\theta_{n-1}^{*}$ denotes model parameters at the end of task $n-1$; $\eta$ denotes the learning rate; and $\lambda$ denotes the regularization constant. We use $v^{(k)}$ to index a vector $v$ at location $k$. To match the empirical setup in which quadratic regularizers generally function, we assume importance scores for regularizing the $n^{\text{th}}$ task (denoted $\alpha_{n-1}$) are calculated during training of tasks $0$ to $n-1$. These scores stay constant as the $n^{\text{th}}$ task proceeds. The total loss for learning the $n^{\text{th}}$ task follows \autoref{eq:loss}.

Our analysis into the mechanics of quadratic regularization is based on a decomposition of the parameter updates. Specifically, computing the gradient of \autoref{eq:loss}, we see the parameter vector, $\theta_{n}$, updates as follows:
\begin{equation}
\label{eq:vec_update}
\theta_{n} \to \theta_{n} - \eta \left(\nabla_{\theta_{n}} L_{T_{n}} + \lambda \alpha_{n-1} \odot \left(\theta_{n} - \theta_{n-1}^{*}\right) \right).
\end{equation}
Rearranging \autoref{eq:vec_update}, we see the $k^{th}$ model parameter updates as follows:
\begin{equation}
\label{eq:param_update}
\theta_{n}^{(k)} \to \underbrace{\left(1 - \eta \lambda \alpha_{n-1}^{(k)}\right) \theta_{n}^{(k)} + \left(\eta \lambda \alpha_{n-1}^{(k)}\right) \theta_{n-1}^{*(k)}}_{\text{Interpolation b/w current and previous values }} \underbrace{- \eta \frac{\partial L_{T_{n}}}{\partial\theta_{n}^{(k)}}.}_{\text{Task Derivative}}
\end{equation}
The above equation illustrates that under quadratic regularization, \emph{a new task is learned by moving model parameters along task-specific derivatives}, while forgetting of previously learned tasks is prevented by \emph{simultaneously performing an interpolation operation between current and previous values of model parameters}. Essentially, the interpolation operation minimizes drift of important model parameters, thus ensuring one can retain performance on previously learned tasks while learning new ones. To understand how mere interpolation of parameters can help mitigate catastrophic forgetting over multiple iterations, we derive the total change in model parameters between the ($n$-1)$^{\text{th}}$ and the $n^{\text{th}}$ tasks. Specifically, if the task-specific gradient for the $j^{\text{th}}$ training iteration is denoted as $g_j$, then unrolling the parameter updates in \autoref{eq:param_update} via recursive substitution for $i$ iterations yields:
\begin{equation}
\label{eq:unrolled}
\theta_{n}^{(k)} = \theta_{n-1}^{*(k)} - \sum_{j = 0}^{i-1} \underbrace{\left[\left(1 - \eta \lambda \alpha_{n-1}^{(k)}\right)^{i-j-1}\eta \right]}_{\text{Effective Learning Rate}} g_{j}^{(k)}.
\end{equation}
\autoref{eq:unrolled} shows that, \emph{based upon its importance, quadratic regularizers change the learning rate of a parameter}. In particular, the effective learning rate at which the $k^{\text{th}}$ parameter updates is calculated by multiplying $\eta$ with exponents of $1 - \eta \lambda \alpha_{n-1}^{(k)}$. If the parameter’s importance is high, the value of $1 - \eta \lambda \alpha_{n-1}^{(k)}$ becomes smaller, consequently reducing the parameter's learning rate and restricting its change. If the parameter’s importance is low, the value of $1 - \eta \lambda \alpha_{n-1}^{(k)}$ remains essentially unchanged; consequently, the parameter's effective learning rate is approximately equal to $\eta$, allowing it to freely change according to task derivatives and enable learning of new tasks.

Overall, \autoref{eq:param_update} and \autoref{eq:unrolled} establish the exact mechanisms by which quadratic regularizers help mitigate catastrophic forgetting. Specifically, \autoref{eq:param_update} shows that in the short-term, quadratic regularizers use a weighted interpolation of model parameters with their previous values to prevent changes to important parameters. Meanwhile, \autoref{eq:unrolled} shows that, in the long-term, quadratic regularizers reduce the learning rate of important parameters, slowing their change as new tasks are learned. These mechanisms alter a model’s optimization path, allowing unimportant parameters to adapt and accommodate new tasks, while restricting movement of important parameters to retain performance on previously learned tasks and prevent catastrophic forgetting. We also note that since the weights used in parameter interpolation (\autoref{eq:param_update}) depend on the product of the learning rate ($\eta$), the regularization constant ($\lambda$), and the parameter’s importance ($\alpha^{(k)}$), essentially these three variables determine the effectiveness of a regularizer.

%% file: files/limitations.tex
\subsection{Limitations in Quadratic Regularization}
\label{sec:limitations}
Having elucidated the role of parameter interpolation in the effectiveness of quadratic regularizers, we now show how the dependence of this operation on the product $\eta \lambda \alpha^{(k)}$ introduces several drawbacks. We intentionally interleave our analysis with experimental evidence to show our claims hold well in practice. 

\setlength{\intextsep}{0pt}%
\begin{wrapfigure}{R}{6cm}
\begin{center}
\includegraphics[width=\linewidth, scale=1]{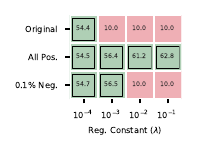}
\end{center}
\caption{\textbf{Case 1: $\eta \lambda \alpha_{n-1}^{(k)} < 0$. Negative importance scores lead to unstable training.} Green, outlined (red) cells imply stable (unstable) training. We analyze a 6-layer CNN trained using SI on 10 CIFAR-100 tasks. \emph{Original} implies both positive and negative scores are allowed, \emph{All Pos.} implies all scores are positive, and \emph{0.1\% Neg.} implies only 0.1\% of parameters are allowed negative scores. Average accuracy is noted inside figure cells (random accuracy=$10\%$, since any given task has 10 classes; plain fine-tuning accuracy=$55.2\%$). Training is always stable when all scores are positive. However, allowing even a few negative scores produces unstable training. A small $\lambda$ mitigates this behavior, but reduces the regularizer's effectiveness, resulting in similar performance to plain fine-tuning. Results on other datasets are provided in the appendix.\vspace{-5pt}}
\label{fig:negmap}
\end{wrapfigure}

\textbf{Setup:} We study four widely used quadratic regularizers: \emph{EWC} (\cite{EWC}), \emph{MAS} (\cite{MAS}), \emph{SI} (\cite{SI}), and \emph{RWalk} (\cite{RWalk}). Training without any strategy to mitigate catastrophic forgetting is called \emph{plain fine-tuning}. We use a 6-layer CNN trained using SGD at a fixed learning rate ($\eta$) of 0.001 on CIFAR-100 tasks (10 tasks; 10 classes per task). This is similar to architectures used in prior work: \cite{SI} use a 6-layer model (4 convolutional+2 dense layers); \cite{RWalk} use a 5-layer model (4 convolutional+1 dense layer); \cite{MAS} use AlexNet (5 convolutional+3 dense layers). We define training as \emph{unstable} if loss proceeds towards infinity, yielding out of precision (\emph{NaN}) gradients and parameters. Our experiments in \autoref{sec:lim1} follow the 1 epoch, batch-size 10 setting advocated by \cite{gem}, who argue that for lifelong learning problems, the model should be allowed to see a sample only once. We highlight that training instabilities elucidated in this section emerge in the first few training iterations themselves ($<$10, generally); hence, these results are not sensitive to the number of epochs. Our analysis in \autoref{sec:lim2} trains model in the stable training regime, where number of epochs can matter. We thus provide results for both 1 epoch, batch-size 10 and 30 epochs, batch-size 256 setting. We find our results follow the same qualitative patterns for both settings. Results are reported on the following metrics (\cite{RWalk}): (a) \emph{average accuracy}, defined as average test accuracy achieved by the final model over all tasks, and (b) \emph{average forgetting}, defined as the average amount of forgetting over all tasks, where an individual task's forgetting is described as the difference between the maximum and final accuracy achieved on it. All results are averaged over five seeds; standard deviations are reported as error bars in figures. For results on other datasets, see appendix.

\subsubsection{Extrapolation and Training Instability}
\label{sec:lim1}
For the $k^{\text{th}}$ parameter, the interpolation operation at every training iteration follows: $\left(1 - \eta \lambda \alpha_{n-1} \right) \odot \mathbf{\theta}_{n}^{(k)} + \left(\eta \lambda \alpha_{n-1} \right) \odot \mathbf{\theta}_{n-1}^{*(k)}$. Since there are generally no constraints on the training hyperparameters and since the scale of importance values can be arbitrarily large or small (e.g., see \autoref{fig:imp_scale}), it is possible for the value of $\eta \lambda \alpha^{(k)}$ to become either negative or greater than 1. In both cases, the interpolation operation converts into an \emph{extrapolation} operation. In this extrapolation regime, we find quadratic regularizers witness unstable training and severe performance degradation. 

\textbf{Case 1.} $\eta \lambda \alpha_{n-1}^{(k)} < 0$.
For parameters with negative importance, the value of $1 - \eta \lambda \alpha_{n-1}^{(k)} > 1$. This pushes the regularizer to the extrapolation regime for such parameters. To understand the implications of this setting, recall the learning rate of a parameter is multiplied by exponents of $1 - \eta \lambda \alpha_{n-1}^{(k)}$ (see \autoref{eq:unrolled}). This implies, for finite gradient, \emph{parameters with negative importance will experience exponentially large updates}, resulting in unstable training. 

To experimentally test this claim, we analyze SI, which allows negative importance scores to be assigned to a parameter. We use SI to train models for different values of $\lambda$ with a fixed learning rate of 0.001. We compare three cases: (a) the original method, which allows both positive and negative scores; (b) using absolute values of the assigned importance scores, thus ensuring all parameters have positive scores; and (c) using absolute values of the importance scores for all but 0.1\% randomly picked, negative importance parameters, hence allowing a very small number of negative scores. The results are shown in \autoref{fig:negmap} and lead to the following observations. (i) When only positive scores are allowed, training remains stable and the final model achieves high average accuracy. (ii) The original method, which allow both positive and negative scores, results in unstable training. In fact, allowing even a few negative scores leads to unstable training. For a very small value of $\lambda$, we note that this behavior is partially mitigated. This can be attributed to the fact the exponent's base, $1 - \eta \lambda \alpha_{n-1}^{(k)}$, remains approximately 1 for small $\lambda$. However, for this setting, the amount of interpolation is very low and hence the effectiveness of quadratic regularization is reduced, resulting in similar performance to plain fine-tuning.

The above experiment corroborates our claim that for negative importance scores, training becomes unstable. This result also shows that \emph{importance scores should be positive for all parameters.} This raises the question, why does SI work well? Interestingly, we find that even though in their paper the authors propose SI as a signed method (i.e., negative scores are allowed) and derive theoretical results for such a signed importance setting, their implementation actually uses a rectifier function (ReLU) to remove negative scores during training (\cite{sigit}). This detail is not discussed in the paper and is likely why the method works well in practice. 

\begin{figure}
\centering
\centerline{\includegraphics[width=\columnwidth]{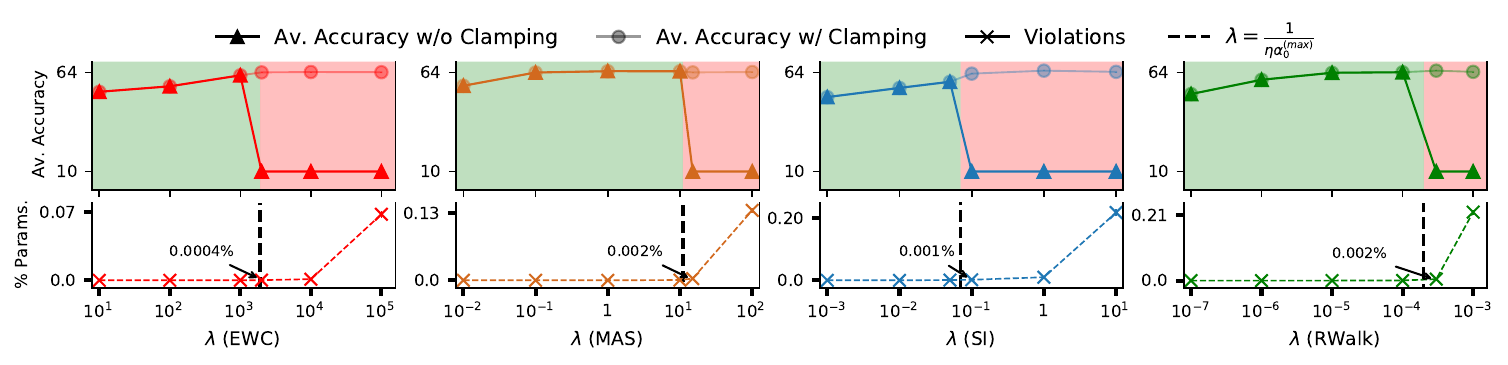}}
\caption{\textbf{Case 2: $\eta \lambda \alpha_{n-1}^{(k)} > 1$. Percentage of parameters that violate $\eta \lambda \alpha_{0}^{(k)} < 1$, leading to unstable training.} We use different values of $\lambda$ to train a 6-layer CNN on CIFAR-100 tasks. We include results with and without clamping of importance scores, which enforces the constraint that $\eta \lambda \alpha_{0}^{(k)} \le 1$ for all parameters. A dotted line marks $\lambda$ at which $\eta \lambda \alpha_{n-1}^{(k)} = 1$ for just one parameter--we first expect to see unstable training right after this value. Green (Red) shading implies we expect training to be stable (unstable) for the corresponding $\lambda$. As shown, across 1.2 million parameters, with even a few violations, training becomes unstable: 5 (0.0004\% parameters) for EWC, 26 (0.002\% parameters) for MAS, 12 (0.001\% parameters) for SI, 31 (0.002\% parameters) for RWalk. The use of clamping helps avoid this instability and yields good performance, indicating the violating parameters were the cause for training instability. Results on other datasets are provided in the appendix.}
\label{fig:violations}
\end{figure}

\textbf{Case 2.} $\eta \lambda \alpha_{n-1}^{(k)} > 1$.
Note that if $\eta \lambda \alpha_{n-1}^{(k)} > 1$ for a parameter, the value of $1 - \eta \lambda \alpha_{n-1}^{(k)} < 0$ for it. That is, similar to Case 1, the regularizer is again pushed into the extrapolation regime. Since $1 - \eta \lambda \alpha_{n-1}^{(k)}$ is multiplied to the current parameter value at every iteration (see \autoref{eq:param_update}), for a reasonably high value of the parameter (i.e., greater than the gradient), we see the parameter's \emph{sign will be flipped (changed) every update}, resulting in unstable training. 

To demonstrate this claim, we train models using different quadratic regularizers and calculate the number of parameters that violate the inequality $\eta \lambda \alpha_{0}^{(k)} < 1$ (i.e., after first task) for various regularization constants. As shown in \autoref{fig:violations}, training is \emph{always} stable when $\lambda$ is small enough to ensure the inequality is not violated. However, with even a \emph{few} violations (\emph{$<$4--30 out of 1.2 million parameters}), training becomes unstable. To confirm that it is indeed these few violating parameters that cause instability, we also include results with a ``clamping'' operation, which enforces the constraint $\eta \lambda \alpha_{0}^{(k)} \le 1$ for all parameters by reassigning violating parameters an importance score of $\nicefrac{1}{\eta\lambda}$. As can be seen in the figure, the clamped models do not witness any instability and are able to achieve high performance. 

Combined, these results corroborate our claim that if $\eta \lambda \alpha_{n-1}^{(k)} > 1$, training becomes unstable. We stress that even though the clamping operation in the experiments above helps address unstable training, its imposed constraint of $\eta \lambda \alpha_{n-1}^{(k)} = 0$ will yield an effective learning rate of $0$ for violating parameters  (see \autoref{eq:vec_update2}). This implies such parameters will be frozen at their current value and not allowed to change. Consequently, if an overly large $\lambda$ is used, a significant portion of the model may remain frozen, disallowing learning of new skills. 

Overall, the above two cases complete our analysis of training instability in the extrapolation regime. Experimentally, we find that as long as the regularizer is in the interpolation regime, training remains stable. In fact, during hyperparameter tuning, we recommend the constraint $0 < \eta \lambda \alpha_{n-1}^{(k)} < 1$ for all parameters should be used to reject values of $\eta$ and $\lambda$ that are likely to produce unstable training.

\subsubsection{Disparate Importance Assignment}
\label{sec:lim2}


\setlength{\intextsep}{0pt}%
\begin{wrapfigure}{R}{6cm}
\begin{center}
\includegraphics[width=\linewidth, scale=1]{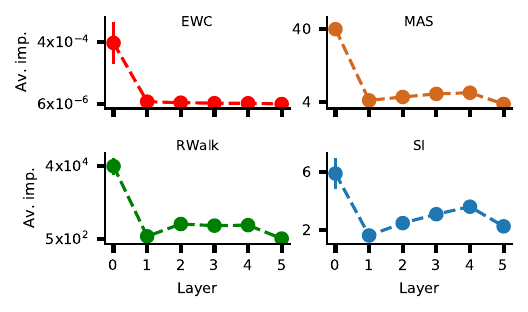}
\end{center}
\caption{\textbf{Average importance is lower for parameters in deeper layers.} We train a 6-layer CNN on the first CIFAR-100 task and analyze importance definitions used by popular quadratic regularizers. As shown, except for SI, importance definitions used in quadratic regularizers result in much lower average importance for parameters in deeper layers. Results on other datasets are provided in the appendix.}
\label{fig:imp_scale}
\end{wrapfigure}

We find that importance definitions used in popular regularizers can assign (much) lower importance to deeper layers than to earlier layers (e.g., see results for EWC in \autoref{fig:imp_scale}). This phenomenon can be explained by borrowing a result from the field of network pruning, where similar importance definitions are used for removing unnecessary parameters in DNNs (\cite{shrinkbench}). Specifically, due to rescale symmetry in DNNs with ReLU non-linearities, layerwise norms of model parameters are conserved across a model (\cite{du, mechanics}). \cite{synflow} recently show that to satisfy this property, generally used importance definitions assign much lower importance to parameters in deeper layers than to parameters in earlier layers. 

\begin{figure}
\begin{subfigure}{.5\textwidth}
  \centering
  \centerline{\includegraphics[width=\columnwidth]{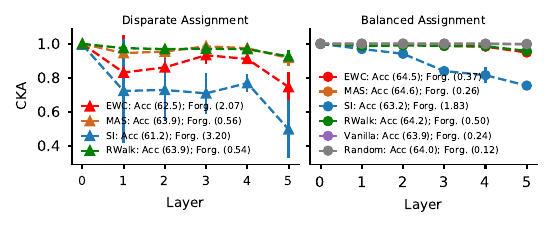}}
  \caption{1 epoch.}
  \label{fig:1epochs}
\end{subfigure}%
\begin{subfigure}{.5\textwidth}
  \centering
  \centerline{\includegraphics[width=\columnwidth]{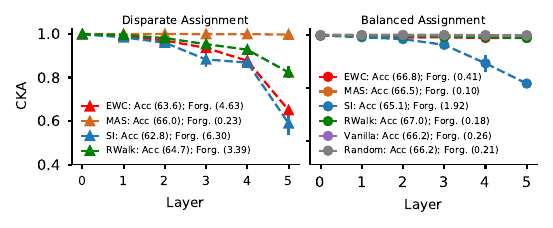}}
  \caption{30 epochs.}
  \label{fig:30epochs}
\end{subfigure}
\caption{\textbf{Balanced importance scores prevent forgetting more effectively.} We use CKA to measure representational similarity between layers of a model trained only on the first task of CIFAR-100 versus the model trained on all tasks. Average accuracy (Acc.) and average forgetting (Forg.) are reported in the legend. Balanced variants ensure importance scores are not biased against deeper layers; they achieve high representational similarity across all layers with minimal forgetting. In contrast, the other measures have significantly lower CKA similarity in deeper layers and, hence, witness more forgetting. Results on other datasets are provided in the appendix.}
\label{fig:cka_analysis}
\end{figure}

To understand the implications of this disparate importance assignment, recall that for stable training, the constraint $\eta \lambda \alpha_{n-1}^{(k)} < 1$ must be satisfied for all parameters. Thus, assuming stable training and a given value of $\eta$, the regularization constant ($\lambda$) will be bounded as follows:
\begin{equation}
\label{eq:lambda_upper}
\lambda < \lambda_{\text{upper}} = \min_{k} \left[\frac{1}{\eta \, \alpha_{n-1}^{(k)}}\right] = \frac{1}{\eta\,\max_{k} \alpha_{n-1}^{(k)}}.
\end{equation}
\autoref{eq:lambda_upper} shows that under stable training, \emph{the valid range for $\lambda$ is restricted by the largest importance score assigned to a model parameter}. However, given the disparate importance assignment in popular quadratic regularizers, will this range be suitable for all parameters in a model? To understand this, we note for a general parameter $\theta_{n}^{l}$, the value of $\eta \lambda \alpha_{n-1}^{(l)}$ is bounded as follows:
\begin{equation}
\label{eq:ratio}
\eta \lambda \alpha_{n-1}^{(l)} < \eta \lambda_{\text{upper}} \alpha_{n-1}^{(l)} = \frac{\alpha_{n-1}^{(l)}}{\max_{k} \alpha_{n-1}^{(k)}}.
\end{equation}
If $\alpha_{n-1}^{(l)} \ll \max_{k} \alpha_{n-1}^{(k)}$, \autoref{eq:ratio} shows the product $\eta \lambda \alpha_{n-1}^{(l)} \to 0$. Thus, \emph{regardless of how large the value of $\lambda$ is}, we find parameters with relatively lower importance will be permitted to change rapidly during training, enabling the model to learn new tasks. Indeed, this is the desired behavior. \emph{However}, in the context of DNNs, we have shown standard importance definitions in quadratic regularizers can assign much lower importance to deeper layers (see \autoref{fig:imp_scale}). This indicates \emph{currently used quadratic regularizers will allow a model to change its deeper layers more readily than its earlier layers}. As shown by \cite{raghu}, forgetting in DNNs is primarily caused by adaptation of deeper layers to recent tasks. Thus, by using lower importance scores for parameters in deeper layers, quadratic regularizers are unable to address the primary source of catastrophic forgetting in DNNs: changes to deeper layers. 

Our analysis above also indicates treating deeper layers on the similar scale of importance as earlier layers should allow a quadratic regularizer to prevent forgetting more easily. To experimentally test this, we design quadratic regularizers with balanced importance assignments across layers. This is achieved via rescaling layerwise importance scores, such that the average importance of scores in any layer is the same. We also consider two naive baselines that satisfy this constraint: (a) \emph{Vanilla}--assign unit importance to all parameters, and (b) \emph{Random}--assign uniformly picked, random importance between $[0, 1]$ to all parameters. Following \cite{raghu}, we analyze each layer's contribution to catastrophic forgetting by using Centered Kernel Alignment (CKA) (\cite{CKA}) and evaluate similarity of feature representations between a model trained only on the first task and a model trained on all tasks. If forgetting is low, features are similar and CKA is high across all layers of the two models. We use the first 3 tasks of our CIFAR-100 setup to perform a hyperparameter search for $\lambda$. The remaining 7 tasks are used for training and evaluation. We train models for both 1 epoch with a batch-size of 10 and 30 epochs with a batch-size of 256. Results are shown in \autoref{fig:cka_analysis}. As can be seen, models trained using balanced quadratic regularizers forget the least and have comparable average accuracy. CKA similarity is high across all layers. In contrast, methods with disparate importance assignment forget more and show low CKA similarity in deeper layers. The above experiment corroborates our claim that \emph{catastrophic forgetting is better prevented by assigning similar importance to all layers, than by using existing, more complex importance definitions.} Surprisingly, prior work on quadratic regularization generally do not evaluate their methods against such simple baselines like Vanilla and Random. \footnote{We note that \cite{EWC} do have a comparison to distance based regularization, which is similar to our Vanilla technique, but their results are for a setting where one computes and stores importance scores computed after every task. This yields a memory complexity that is proportional to the product of model size and number of tasks, ignoring one of the core desiderata of continual learning that memory requirements should grow at most sub-linearly with the number of tasks (\cite{thesis}).}

%% file: files/explicitreg.tex
\section{Addressing Limitations in Quadratic Regularization}
\label{sec:emr}

Having established the mechanisms by which quadratic regularizers prevent catastrophic forgetting and the corresponding drawbacks of those mechanisms, we seek a simple way to avoid these problems and still prevent catastrophic forgetting. To this end, we note our analysis in \autoref{sec:limitations} shows that two primary conditions should be satisfied to ensure a quadratic regularizer performs well: (a) the regularizer should function in the interpolation regime and (b) importance scores be relatively constant across layers. We show these conditions can be easily satisfied by breaking the quadratic regularizer update (\autoref{eq:param_update}) into two operations:
\begin{equation} 
\label{eq:emr}
\begin{split}
&\text{(i)}\, \theta_{n} \to \theta_{n} - \eta \nabla_{\theta_{n}} L_{T_{n}} \quad\quad\quad \text{{\small (Task-specific change) \text{ and}}}\\
&\text{(ii)}\, \theta_{n} \to \left(1-R_{j}\right) \odot \theta_{n} + R_{j} \odot \theta_{n-1}^{*} \quad \text{{\small (Interpolation)}}.\\
\end{split}
\end{equation}
Here, the vector $R_{j}$ is defined to control the amount of interpolation at iteration $j$. Contrasting this update with the quadratic regularization update in \autoref{eq:param_update}, one can see the main difference lies in the order of operations: while quadratic regularizers perform the task-specific update and the interpolation operation \emph{in a single step}, our proposed modification breaks that update into \emph{two separate operations}. This makes the interpolation operation \emph{explicit}, instead of the implicit operation in quadratic regularization. .

To understand the benefit of making interpolation an explicit operation, notice that in \autoref{eq:emr}, \emph{the amount of interpolation is independent of the training hyperparameters}; only the vector $R_{j}$ controls the amount of interpolation. This allows us to circumvent the need of a regularization coefficient and helps avoid training instabilities arising from gradient descent dynamics (see \autoref{sec:lim1}). Further, we can exploit this independence to define $R_{j}$ in a manner such that the desired conditions for effective quadratic regularization are always satisfied. To this end, we define $R_{j}$ as a measure of the importance of a parameter for previous tasks relative to its importance for all tasks. Specifically, assume learning of the $n^{\text{th}}$ task is currently underway. Again using $\alpha_{n-1}$ to denote importance of parameters according to tasks $0$ to $n-1$ and $\alpha_{T_{n}}$ to denote the importance of parameters according to the $n^{\text{th}}$ task, we define the relative importance of a parameter $\theta_{n}^{(k)}$ as
\begin{equation} 
\label{eq:imp_ratios}
R_{j}^{(k)} = \frac{\sqrt{\alpha_{n-1}^{(k)}}}{\sqrt{\alpha_{T_{n}}^{(k)}} + \sqrt{\alpha_{n-1}^{(k)}}}.
\end{equation}
Here, the definition of importance of a parameter can be borrowed from any popular quadratic regularizer. Note that under this formulation, $R_{j}^{(k)}$ necessarily ranges from $0$ to $1$ for all parameters, hence satisfying condition (a) and ensuring stable training. If $R_{j}^{(k)}$ is large (closer to 1), then the $k^{\text{th}}$ parameter is relatively more important for preserving performance on previous tasks than for the current task. In this case, the parameter's \emph{previous} value is weighted more heavily to prevent its change. If $R_{j}^{(k)}$ is small (closer to 0), then the $k^{\text{th}}$ parameter is relatively less important for previous tasks. In this case, the parameter's \emph{current} value is weighted more heavily, allowing the parameter to adapt to a new task. Finally, we note that even if a parameter's absolute importance is low, its relative importance can be high if it is unimportant to previous tasks. \emph{Thus, using this definition for $R_{j}$, we can also avoid problems related to assignment of lower importance scores to deeper layers.}

We also add that the use of square root in \autoref{eq:imp_ratios} yields us small-but-significant improvements over use of simple magnitude ($\sim1$--$2$\% less forgetting, in general). To understand why, recall importance scores in our paper are computed in a running average manner. In a small-batch scenario (one of our primary setups), gradient noise is very high, especially in the early iterations (magnitude of order 100). This noise manifests as gradient spikes that produce large fluctuations in the current task's importance scores. Since the previously computed importance scores remain fixed and since we regularize the model using relative importance to previous tasks' importance scores, these gradient spikes bias learning towards the current task. To avoid this behavior, we use a sublinear function (square root) to minimize sensitivity to fluctuating scores. In fact, we found similar improvements with cube root too (another sublinear function). Over a few iterations, gradient noise is removed by the running average, thus improving signal to noise ratio and yielding reliable estimates. Thus, in the later iterations, both square root and magnitude show similar behavior. However, square root has better ability to avoid forgetting in the early iterations and hence its overall forgetting ends up being smaller.

\begin{algorithm}[tb]
   \caption{Quadratic Regularization with Explicit Interpolation Steps}
   \label{alg:emr}
\begin{algorithmic}
   \STATE {\bfseries Input:} parameterization $\theta_{n-1}^{*}$ and importance $\alpha_{n-1}$ after learning $n-1$ tasks; number of training iterations \#iters and Loss $L_{T_{n}}$ for learning the $n^{\text{th}}$ task.
   \STATE {\bfseries Initialize:} $\theta_{n} = \theta_{n-1}^{*}$; $\alpha_{T_{n}} = \mathbf{0}$.
   \FOR{$j=0$ {\bfseries to} \#iters-1}
   \STATE $\theta_{n} \to \theta_{n} - \eta \nabla_{\theta_{n}} L_{T_{n}}$ \hfill{\small (Task-specific change)}
   \STATE Update $\alpha_{T_{n}}$ 
   \STATE Compute $R_{j} = \frac{\sqrt{\alpha_{n-1}}}{\sqrt{\alpha_{T_{n}}} + \sqrt{\alpha_{n-1}}}$ \hfill{\small (Relative importance)}              
   \STATE $\theta_{n} \to \left(1-R_{j}\right) \odot \theta_{n} + \left(R_{j}\right) \odot \theta_{n-1}^{*}$ \hfill{\small (Interpolation)}
   \ENDFOR
    \STATE $\alpha_{n} \to (\alpha_{T_{n}} + \alpha_{n-1})/2$ \hfill{\small (Update importance)}
   \STATE {\bfseries Return:} $\theta_{n}$, $\alpha_{n}$
\end{algorithmic}
\end{algorithm}

The overall algorithm for quadratic regularization with explicit interpolation steps is shown in \autoref{alg:emr}. Before proceeding, we highlight that instead of using our explicit update formulation, the dependence on hyperparameters could also have been removed by scaling down the regularization coefficient in standard quadratic regularizers with the model learning rate. However, this would still retain the presence of hyperparameters in the overall setup, in contrast with our explicit update formulation.

\input{files/emr_vs_qreg}

%% file: files/emr_vs_qreg.tex
\begin{table}[t]
\caption{\label{tab:emr_vs_qreg}Comparison of Average Accuracy (Acc; $\uparrow$ indicates higher is better) and Average Forgetting (Forg; $\downarrow$ indicates lower is better) for plain and explicit interpolation variants of EWC, MAS, SI, and RWalk, averaged over five seeds (std.\ dev.\ are reported in appendix). Results are also provided with two replay based methods (A-GEM and ER-Reservoir) and plain fine-tuning (Plain). We consider three datasets: CIFAR-100 (10 tasks); Oxford-Flowers (17 tasks); and Caltech-256 (32 tasks). For a specific regularizer, the better performing variant is underlined; the best results are in bold. As shown, variants with explicit interpolation steps consistently outperform their Quadratic Regularization counterparts. This behavior is most prominent in datasets with unbalanced classes and longer task sequences (Oxford-Flowers and Caltech-256), where hyperparameter tuning is difficult.}
\tablefont
\begin{subtable}{\textwidth}
\centering
\begin{tabular}{@{}c!{\vrule width 1pt}c!{\vrule width 1pt}cccc|cc!{\vrule width 1pt}cccc!{\vrule width 1pt}cc@{}}
\toprule
\multicolumn{2}{c!{\vrule width 1pt}}{1 Epoch}     & \multicolumn{6}{c!{\vrule width 1pt}}{Plain Quadratic Regularizers}                                 & \multicolumn{4}{c!{\vrule width 1pt}}{Explicit Interpolation Variants} & \multicolumn{2}{c}{Replay} \\ \midrule
CIFAR   & Plain & EWC  & MAS  & SI   & RWalk                  & Van.\ & Rand.\  & EWC        & MAS                 & SI         & RWalk      & A-GEM & ER       \\ \midrule
Acc ($\uparrow$)    & 55.3  & 62.5 & 63.9 & 61.2 & {\ul 63.9} & 63.9 & 64.0 & {\ul 63.8} & {\ul 64.0} & {\ul 63.9} & 63.8       & 67.2 & \textbf{68.8}      \\
Forg ($\downarrow$)   & 9.13  & 2.07 & 0.56 & 3.20 & 0.54     & 0.24 & 0.12 & {\ul 0.08} & {\ul \textbf{0.06}} & {\ul 0.07} & {\ul 0.09} & 1.29 & 0.69 \\ \midrule
Flowers & Plain & EWC  & MAS  & SI   & RWalk      & Van.\       & Rand.\ & EWC & MAS                 & SI         & RWalk      & A-GEM & ER      \\ \midrule
Acc ($\uparrow$)    & 37.9  & 41.3 & 57.5 & 36.2 & 44.9       & 59.5 & 58.5 & {\ul 60.0} & {\ul 59.6}          & {\ul 59.9} & {\ul 59.4} & 68.61 & \textbf{68.9} \\
Forg ($\downarrow$)   & 13.4  & 10.6 & 3.61 & 15.1 & 9.68     & 1.19 & 3.61 & {\ul 1.03} & {\ul \textbf{0.44}}          & {\ul 0.99} & {\ul 1.31} & 4.08 & 3.45\\ \midrule
Cal-256 & Plain & EWC  & MAS  & SI   & RWalk      & Van.\       & Rand.\ & EWC   & MAS                 & SI         & RWalk      & A-GEM & ER      \\ \midrule
Acc ($\uparrow$)    & 40.1  & 41.9 & 53.7 & 40.9 & 43.7       & 55.1 & 56.3 & {\ul 56.2} & {\ul 57.3} & {\ul 56.0} & {\ul 55.8} & 63.9 & \textbf{64.3}\\
Forg ($\downarrow$)   & 6.21  & 5.59 & 3.92 & 5.68 & 4.69     & 1.89 & 1.59 & {\ul 1.41} & {\ul \textbf{0.36}} & {\ul 1.42} & {\ul 1.31} & 2.89 & 2.75 \\ \bottomrule
\end{tabular}
\end{subtable}

\begin{subtable}{\textwidth}
\centering
\begin{tabular}{@{}c!{\vrule width 1pt}c!{\vrule width 1pt}cccc|cc!{\vrule width 1pt}cccc!{\vrule width 1pt}cc@{}}
\toprule
\multicolumn{2}{c!{\vrule width 1pt}}{30 Epochs}       & \multicolumn{6}{c!{\vrule width 1pt}}{Plain Quadratic Regularizers}                                 & \multicolumn{4}{c!{\vrule width 1pt}}{Explicit Interpolation Variants} & \multicolumn{2}{c}{Replay} \\ \midrule
CIFAR   				& Plain & EWC  & MAS  & SI   & RWalk    & Van.\ & Rand.\  		& EWC   & MAS   & SI          & RWalk & A-GEM & ER   \\ \midrule
Acc ($\uparrow$)    	& 60.7  & 63.6 & 66.3 & 62.8 & 64.7 	& 66.2  & 66.2 			& {\ul 66.3} 	& {\ul 66.0}  & {\ul 66.2} & {\ul 66.1}  & 66.4  & \textbf{66.8} \\
Forg ($\downarrow$)   	& 8.01  & 4.63 & 0.23 & 6.30 & 3.39     & 0.26  & 0.21 			& {\bf \ul 0.21} 	& {\ul 0.23}  & {\ul 0.25} & {\ul 0.29} & 3.10  & 1.64 \\ \midrule
Flowers 				& Plain & EWC  & MAS  & SI   & RWalk    & Van.\ & Rand.\ 		& EWC 	& MAS   & SI          & RWalk & A-GEM & ER   \\ \midrule
Acc ($\uparrow$)    	& 49.9  & 54.1 & 60.5 & 51.1 & 55.7     & 60.3  & 60.9 			& {\ul 61.3} 	& {\ul 61.6}  & {\ul 61.8} & {\ul 61.7}  & 68.9 & \textbf{69.4} \\
Forg ($\downarrow$)   	& 11.8  & 9.07 & 1.19 & 12.4 & 7.19     & \textbf{0.36}  & 1.79 			& {\ul 1.07} 	& {\ul 0.48}  & {\ul 0.62} & {\ul 0.71}  & 4.29  & 3.01 \\ \midrule
Cal-256 				& Plain & EWC  & MAS  & SI   & RWalk    & Van.\ & Rand.\ 		& EWC   & MAS   & SI          & RWalk & A-GEM & ER   \\ \midrule
Acc ($\uparrow$)    	& 40.6  & 50.3 & 60.1 & 50.6 & 53.7     & 60.0  & 60.1 			& {\ul 60.6} 	& {\ul 60.9}  & {\ul 60.8} & {\ul 60.6}  & 65.7  & \textbf{66.1} \\
Forg ($\downarrow$)   	& 15.8  & 7.15 & 0.99 & 7.41 & 4.69     & 0.32  & 0.22 			& {\ul 0.22} 	& {\ul \bf 0.18}  & {\ul 0.32} & {\ul 0.34}  & 2.86  & 2.32 \\ \bottomrule
\end{tabular}
\end{subtable}
\end{table}

%% file: files/experiments.tex
\subsection{Experimental Evaluation}
\label{sec:exptt_eval}
We now evaluate the effectiveness of the proposed quadratic regularizers with explicit interpolation steps. We also provide comparisons with two experience replay-based methods: A-GEM (\cite{agem}) and ER-Reservoir (\cite{er_reservoir}). We stress that our objective in reporting this comparison is \emph{not} to beat the state-of-the-art, but to evaluate if our modified quadratic regularizers can reduce the gap between regularization and replay based methods, which can be inappropriate for applications where long-term storage of training data undermines privacy or greatly increases cost (e.g., in embedded systems).

We use three datasets of different complexities: (i) CIFAR-100, divided into 10 tasks with 10 classes per task and 500 samples per class; (ii) Oxford-Flowers, divided into 17 tasks with 6 classes per task and 72 samples per class (on average); and (iii) Caltech-256, divided into 32 tasks with 8 classes per task and 95 samples per class (on average). Note that Oxford-Flowers and Caltech-256 have unbalanced classes and few samples per class, which are characteristics of several real-world applications. We use a 6-layer CNN for all datasets and train the model using SGD with momentum for all tasks; results for ResNet-18 are in the appendix. The first 3 tasks for all datasets are reserved to run a grid search for finding the optimal hyperparameter configuration (learning rate/regularization constant). Since the models are trained in the stable training regime (enforced via hyperparameter tuning), we train models for both 1 epoch, batch-size 10 setting advocated by \cite{gem} and 30 epochs, batch-size 256 setting where we find all models achieve 95--100\% training accuracy when training on a given task, ensuring convergence (an assumption made by prior work to derive their regularizer strategies). Further, following prior works (\cite{MAS, EWC, SI}), we use a multi-head setting. 

Results are shown in \autoref{tab:emr_vs_qreg}. We note explicit interpolation variants \emph{significantly} outperform plain quadratic regularizers. For ex., in the 1 epoch setting, \emph{accuracy improves by 6.2\% on average and up to 23\%.} We further make the following observations: (i) On unbalanced datasets and long task sequences, explicit interpolation results in substantial improvements. However, on CIFAR-100, whose tasks correspond to hundreds of parameter updates, plain quadratic regularizers perform competitively. This indicates when longer task sequences are used for hyperparameter search, the resulting configuration is able to perform well for subsequent tasks as well. In contrast, on Oxford-Flowers and Caltech-256, especially in the 1 epoch setting, where the respective tasks have only 43 and 76 parameter updates (on average), searching for adequate hyperparameters is difficult and plain quadratic regularizers result in up to 10\% forgetting. (ii) With our proposed modifications, quadratic regularizers achieve witness lower forgetting than replay based methods, but continue to underperform on average accuracy. However, we note that when long-term storage of training data is prevented by privacy or cost concerns, our modifications allow quadratic regularizers to serve as a useful and reliable alternative to replay based methods.

%% file: files/conclusion.tex
\section{Conclusion}
In this work, we determine the mechanisms using which quadratic regularization based methods prevent catastrophic forgetting in DNNs. Our primary findings show that in the short-term, these methods interpolate current and past values of model parameters, disallowing change. In the long-term, they change the learning rate of a parameter in proportion to its importance, reducing the overall movement of more important parameters. Our analysis also shows two primary pitfalls that limit the effectiveness of quadratic regularization: (i) extrapolation, instead of interpolation, of parameters due to inappropriate hyperparameters and (ii) lower importance assignment to deeper layers. We propose an explicit interpolation variant of quadratic regularizers, which is able to circumvent these pitfalls, boasting much better performance, consistently. In future work, we aim to perform similar analyses of the mechanisms which allow replay-based techniques to prevent catastrophic forgetting.

\section*{Acknowledgements}
The authors thank Hidenori Tanaka and anonymous reviewers for valuable feedback on the paper. This work was supported in part by NSF under award CNS-2008151.

%% file: appendix.tex
\newpage
\appendix
\section{Appendix}
The appendix is organized as follows:
\begin{itemize}
	\item \autoref{app:related} contains a more detailed related work on strategies to prevent catastrophic forgetting.
	\item \autoref{sec:sec3} provides derivation for the relationship between parameters of $n^{\text{th}}$ and $(n-1)^{\text{th}}$ task (Equation~5 from main paper). 
	\item \autoref{sec:setup_appendix} explains our experimental setup, including details on the datasets used in our experiments, the model architecture, dataset preprocessing, training/evaluation protocol, and the hyperparameter tuning protocol.
	\item \autoref{sec:sec4} provides more results for experiments conducted in \autoref{sec:limitations} of main paper. 
		\subitem \autoref{sec41}: Extrapolation Regime Case 1: $\eta \lambda \alpha_{n-1}^{(k)} < 0$.
		\subitem \autoref{sec42}: Extrapolation Regime Case 2: $\eta \lambda \alpha_{n-1}^{(k)} > 1$.
		\subitem \autoref{sec43}: Disparate Importance Assignment.
	\item \autoref{sec:sec5} provides more results for experiments conducted in \autoref{sec:exptt_eval} of main paper. 
		\subitem \autoref{sec51}: Compares plain and explicit interpolation quadratic regularization variants on ResNet-18.
		\subitem \autoref{sec52}: Contains a detailed version of \autoref{tab:emr_vs_qreg} with standard deviations of the results.
\end{itemize}

\section{Related Work}
\label{app:related}
\textbf{Regularization Based Methods:} We note that beyond quadratic regularization, prior works have also described \emph{functional regularization} strategies based on knowledge distillation (\cite{lwf}) and Bayesian modeling (\cite{funcreg}). In this work, we specifically focus on quadratic regularization techniques.

\textbf{Replay Based Methods:} Another successful approach to mitigate catastrophic forgetting is based on the idea of \emph{experience replay} in biological systems. Such methods maintain a memory buffer of samples from previous tasks and fine-tune the model on this buffer while learning new tasks (\cite{er_reservoir, mer, er_grad}). \cite{dgr} train a \emph{generative model} to learn previous tasks' data distributions and generate synthetic samples for experience replay while learning new tasks. Recent works have also used a \emph{gradient episodic memory} (\cite{gem, agem}) and \emph{orthogonal gradient updates} (\cite{ogd, gpm}) to avoid catastrophic forgetting.

\textbf{Expansion Based Methods:} To mitigate catastrophic forgetting, prior works have also proposed to increase model capacity by allocating more neurons. For example, \cite{PNN} add a small network to the original model every time a new task is learned. The original model remains fixed and the added network is fine-tuned. Similarly, \cite{cl_nas} optimize the model architecture via neural architecture search. This ``optimized'' architecture is then fine-tuned for the next task. \cite{den} design dynamically expandable networks, which split neurons on the arrival of new tasks to increase model capacity. While such methods do not suffer from catastrophic forgetting, they can significantly increase memory consumption, making them infeasible for scenarios with long sequences of tasks.

\input{supplementary/files/sec3}
\input{supplementary/files/setup}
\input{supplementary/files/sec4}

%% file: supplementary/files/sec3.tex
\section{Derivation of relationship between parameters of $n^{\text{th}}$ and $(n-1)^{\text{th}}$ task}
\label{sec:sec3}
\textbf{Preliminaries:} 
We restate notations used throughout the paper for ease of understanding. $\mathbf{\theta}_{n}$ denotes model parameters and $L_{T_{n}}$ denotes a task-specific loss for the $n^{\text{th}}$ task; $\mathbf{\theta}_{n-1}^{*}$ denotes model parameters at the end of task $n-1$; $\mathbf{\theta}_{n}[i]$ denotes model parameterization at the $i^{\text{th}}$ iteration; $\eta$ denotes the overall model learning rate; and $\lambda$ denotes the regularization constant. We use $v^{(k)}$ to index a vector $v$ at location $k$. To match the empirical setup in which quadratic regularizers generally function, we assume importance scores for regularizing the $n^{\text{th}}$ task (denoted $\alpha_{n-1}$) are calculated during training of tasks 0 to $n-1$. These scores stay constant as the $n^{\text{th}}$ task proceeds.

\textbf{Derivation:} \autoref{eq:unrolled} of the main paper describes the relationship between model parameters before and after training for the $n^{\text{th}}$ task. Specifically, the following relationship is noted:
\begin{equation}
\label{eq:unrolled2}
\theta_{n}^{(k)} = \theta_{n-1}^{*(k)} - \sum_{j = 0}^{i-1} \underbrace{\left[\left(1 - \eta \lambda \alpha_{n-1}^{(k)}\right)^{i-j-1}\eta \right]}_{\text{Effective Learning Rate}} g_{j}^{(k)}.
\end{equation}

To derive the above result, we can unroll the model updates under quadratic regularization. Recall, the model updates as follows (\autoref{eq:vec_update} from main paper):
\begin{equation}
\label{eq:vec_update2}
\begin{split}
\mathbf{\theta}_{n}[i+1] &= \mathbf{\theta}_{n}[i] - \eta \left(\nabla_{\mathbf{\theta}_{n}[i]} L_{T_{n}} + \lambda \alpha_{n-1} \odot \left(\mathbf{\theta}_{n}[i] - \mathbf{\theta}_{n-1}^{*}\right) \right). \\
\text{After rearranging:}\quad \mathbf{\theta}_{n}[i+1] &= \left(\mathbf{1} - \eta \lambda \alpha_{n-1}\right) \odot \theta_{n}[i] + \left(\eta \lambda \alpha_{n-1}\right) \odot \theta_{n-1}^{*} - \eta \nabla_{\theta_{n}[i]} L_{T_{n}}.
\end{split}
\end{equation}

Recall, the gradient of task-specific loss at iteration $j$ is denoted as $g_{j} = \nabla_{\theta_{n}[j]} L_{T_{n}}$. Then, \autoref{eq:vec_update2} can be used to show the following:
\begin{equation}
\label{eq:iter1}
\begin{split}
\theta_{n}[1] &= \left(\mathbf{1} - \eta \lambda \alpha_{n-1}\right) \odot \theta_{n}[0] + \left(\eta \lambda \alpha_{n-1}\right) \odot \theta_{n-1}^{*} - \eta g_{0} \\
              &= \left(\mathbf{1} - \cancel{\eta \lambda \alpha_{n-1}}\right) \odot \theta_{n-1}^{*} + \left(\cancel{\eta \lambda \alpha_{n-1}}\right) \odot \theta_{n-1}^{*} - \eta g_{0} \\
              &= \theta_{n-1}^{*} - \eta g_{0} \\
\implies \theta_{n}[1] &= \theta_{n-1}^{*} - \sum_{j = 0}^{0} \left[\left(1 - \eta \lambda \alpha_{n-1}\right)^{1-j-1}\eta \right] \odot g_{j}.
\end{split}
\end{equation}
where the second equality follows from the fact $\theta_{n}[0] = \theta_{n-1}^{*}$ (the model for $n^{\text{th}}$ task is initialized according to $(n-1)^{\text{th}}$ task's final parameterization).

Unrolling for further iterations, we have:
\begin{equation}
\label{eq:iter2}
\begin{split}
\theta_{n}[2] &= \left(\mathbf{1} - \eta \lambda \alpha_{n-1}\right) \odot \theta_{n}[1] + \left(\eta \lambda \alpha_{n-1}\right) \odot \theta_{n-1}^{*} - \eta g_{1} \\
              &= \left(\mathbf{1} - \eta \lambda \alpha_{n-1}\right) \odot \left(\theta_{n-1}^{*} - \eta g_{0}\right) + \left(\eta \lambda \alpha_{n-1}\right) \odot \theta_{n-1}^{*} - \eta g_{1} \\
              &= \left(\mathbf{1} - \cancel{\eta \lambda \alpha_{n-1}}\right) \odot \theta_{n-1}^{*} + \left(\cancel{\eta \lambda \alpha_{n-1}}\right) \odot \theta_{n-1}^{*} - \eta\left(\mathbf{1} - \eta \lambda \alpha_{n-1}\right) \odot g_{0}  - \eta g_{1} \\
              &= \theta_{n-1}^{*} - \eta\left(\mathbf{1} - \eta \lambda \alpha_{n-1}\right) \odot g_{0} - \eta g_{1} \\
\implies \theta_{n}[2] &= \theta_{n-1}^{*} - \sum_{j = 0}^{1} \left[\left(1 - \eta \lambda \alpha_{n-1}\right)^{2-j-1}\eta \right] \odot g_{j}.
\end{split}
\end{equation}

Assume, for the $i^{\text{th}}$ iteration, following holds true:
\begin{equation}
\label{eq:iteri}
\theta_{n}[i] = \theta_{n-1}^{*} - \sum_{j = 0}^{i-1} \left[\left(1 - \eta \lambda \alpha_{n-1}\right)^{i-j-1}\eta \right] \odot g_{j}.
\end{equation}

Then, for the $(i+1)^{\text{th}}$ iteration, we have:
\begin{equation}
\label{eq:iteriplus1}
\begin{split}
\theta_{n}[i+1] &= \left(\mathbf{1} - \eta \lambda \alpha_{n-1}\right) \odot \theta_{n}[i] + \left(\eta \lambda \alpha_{n-1}\right) \odot \theta_{n-1}^{*} - \eta g_{i} \\
              &= \left(\mathbf{1} - \eta \lambda \alpha_{n-1}\right) \odot \left(\theta_{n-1}^{*} - \sum_{j = 0}^{i-1} \left[\left(1 - \eta \lambda \alpha_{n-1}\right)^{i-j-1}\eta \right] \odot g_{j}\right) + \left(\eta \lambda \alpha_{n-1}\right) \odot \theta_{n-1}^{*} - \eta g_{i} \\
              &= \left(\mathbf{1} - \cancel{\eta \lambda \alpha_{n-1}}\right) \odot \theta_{n-1}^{*} + \left(\cancel{\eta \lambda \alpha_{n-1}}\right) \odot \theta_{n-1}^{*} - \sum_{j = 0}^{i-1} \left[\left(1 - \eta \lambda \alpha_{n-1}\right)^{i+1-j-1}\eta \right] \odot g_{j}  - \eta g_{i} \\
              &= \theta_{n-1}^{*}  - \sum_{j = 0}^{i-1} \left[\left(1 - \eta \lambda \alpha_{n-1}\right)^{i+1-j-1}\eta \right] \odot g_{j}  - \eta g_{i}  \\
\implies \theta_{n}[i+1] &= \theta_{n-1}^{*} - \sum_{j = 0}^{i} \left[\left(1 - \eta \lambda \alpha_{n-1}\right)^{i+1-j-1}\eta \right] \odot g_{j}.
\end{split}
\end{equation}

Hence, assuming \autoref{eq:iteri} holds true for the $i^{\text{th}}$ iteration, we see it holds true for the $(i+1)^{\text{th}}$ iteration as well. Since the relationship is true for $i=1$ (see \autoref{eq:iter1}) and $i=2$ (see \autoref{eq:iter2}), thus, by principle of induction, the relationship holds true for all $i$. This completes the derivation.

%% file: supplementary/files/setup.tex
\section{Experimental Setup}
\label{sec:setup_appendix}
We provide further details on our experimental setup in this section. 

\subsection{Datasets}
We use CIFAR-100, Oxford-Flowers, and Caltech-256 datasets in our work.

\textbf{CIFAR-100:} CIFAR-100 is a standardly used datasets for evaluating continual/lifelong learning algorithms and has 100 classes overall. We divide the dataset into 10 tasks for all experiments in the paper. Each task corresponds to 10 sequential classes. Any given class has 500 train samples and 100 test samples. The dataset is available at \url{https://www.cs.toronto.edu/~kriz/cifar.html}.

\textbf{Oxford-Flowers:} Oxford-Flowers is a collection of 102 classes of flowers and is divided into a train/val/test split by default. We divide the dataset into 17 tasks and 6 classes per task. Similar to CIFAR-100, we use sequential classes for Oxford-Flowers as well. Designed for few-shot learning, the dataset has very few training samples per class. We thus use both the train and validation splits to perform training, while the test split is used for testing. The dataset is unbalanced, with different number of samples for different classes. On average, each class has 72 training and 20 test samples. The dataset is available at \url{https://www.robots.ox.ac.uk/~vgg/data/flowers/}.

\textbf{Caltech-256:} Caltech-256 is a standard classification benchmark with 256 classes. We divide the dataset into 32 tasks with 8 classes per task. Similar to CIFAR-100, we use sequential classes for Caltech-256 as well. The dataset is unbalanced, with different number of samples for different classes. On average, each class has 95 training and 27 test samples. The dataset is available at \url{http://www.vision.caltech.edu/Image_Datasets/Caltech256/}.

\begin{table}
\scriptsize
\centering
\caption{Hyperparamaters found using grid search for different methods/datasets for the 1 epoch, 10 batch-size setting.}
\label{tab:params}
\begin{tabular}{@{}c|cccc|cc|cccc|cc@{}}
\toprule
Dataset & \multicolumn{6}{c|}{Plain Quadratic Regularizers}                        & \multicolumn{4}{c}{Explicit Interpolation Variants} & \multicolumn{2}{c}{Replay-Based} \\ \midrule
CIFAR   & EWC   & MAS    & SI    & RWalk   & Van.  & Rand  & EWC   & MAS   & SI    & RWalk  & A-GEM & ER\\ \midrule
LR ($\eta$)      & 0.001 & 0.001  & 0.001 & 0.003   & 0.001 & 0.001 & 0.001 & 0.001 & 0.001 & 0.001 & 0.001 & 0.001 \\
Reg ($\lambda$)     & $10^{3}$  & $10^{-2}$   & $10^{-2}$  & $10^{-4}$  & $10^{2}$   & $10^{2}$   & -     & -     & -     & -     & -     & - \\ \midrule
Flowers & EWC   & MAS    & SI    & RWalk   & Van.  & Rand  & EWC   & MAS   & SI    & RWalk & A-GEM & ER \\ \midrule
LR ($\eta$)      & 0.003 & 0.003  & 0.003 & 0.003   & 0.003 & 0.003 & 0.003 & 0.003 & 0.003 & 0.003 & 0.003 & 0.003 \\
Reg ($\lambda$)     & $10^{2}$   & $10^{-1}$    & $10^{-3}$ & $10^{-5}$ & $10^{3}$  & $10^{1}$    & -     & -     & -     & -     & -     & -     \\ \midrule
Cal-256 & EWC   & MAS    & SI    & RWalk   & Van.  & Rand  & EWC   & MAS   & SI    & RWalk & A-GEM & ER \\ \midrule
LR ($\eta$)      & 0.001 & 0.001  & 0.001 & 0.001   & 0.003 & 0.003 & 0.001 & 0.001 & 0.001 & 0.001 & 0.003 & 0.003 \\
Reg ($\lambda$)     & $10^{3}$  & $10^{-1}$ & $10^{-4}$   & $10^{-5}$ & $10^{2}$   & $10^{2}$     & -     & -     & -     & -     & -     & -     \\ \bottomrule
\end{tabular}
\end{table}

\begin{table}
\scriptsize
\centering
\caption{Hyperparamaters found using grid search for different methods/datasets for 30 epochs, 256 batch-size setting.}
\label{tab:params_30 epochs}
\begin{tabular}{@{}c|cccc|cc|cccc|cc@{}}
\toprule
Dataset & \multicolumn{6}{c|}{Plain Quadratic Regularizers}                        & \multicolumn{4}{c}{Explicit Interpolation Variants} & \multicolumn{2}{c}{Replay-Based} \\ \midrule
CIFAR   & EWC   & MAS    & SI    & RWalk   & Van.  & Rand  & EWC   & MAS   & SI    & RWalk  & A-GEM & ER\\ \midrule
LR ($\eta$)      & 0.001 & 0.001  & 0.001 & 0.003   & 0.001 & 0.001 & 0.001 & 0.001 & 0.001 & 0.001 & 0.001 & 0.001 \\
Reg ($\lambda$)     & $10^{3}$  & $10^{-2}$   & $10^{-1}$  & $10^{-3}$  & $10^{2}$   & $10^{2}$   & -     & -     & -     & -     & -     & - \\ \midrule
Flowers & EWC   & MAS    & SI    & RWalk   & Van.  & Rand  & EWC   & MAS   & SI    & RWalk & A-GEM & ER \\ \midrule
LR ($\eta$)      & 0.003 & 0.003  & 0.003 & 0.003   & 0.003 & 0.003 & 0.003 & 0.003 & 0.003 & 0.003 & 0.003 & 0.003 \\
Reg ($\lambda$)     & $10^{2}$   & $10^{-2}$    & $10^{-1}$ & $10^{-3}$ & $10^{3}$  & $10^{1}$    & -     & -     & -     & -     & -     & -     \\ \midrule
Cal-256 & EWC   & MAS    & SI    & RWalk   & Van.  & Rand  & EWC   & MAS   & SI    & RWalk & A-GEM & ER \\ \midrule
LR ($\eta$)      & 0.001 & 0.001  & 0.001 & 0.001   & 0.003 & 0.003 & 0.001 & 0.001 & 0.001 & 0.001 & 0.003 & 0.003 \\
Reg ($\lambda$)     & $10^{3}$  & $10^{-2}$ & $10^{-2}$   & $10^{-4}$ & $10^{2}$   & $10^{2}$     & -     & -     & -     & -     & -     & -     \\ \bottomrule
\end{tabular}
\end{table}

\subsection{Training Setup}
In this section, we report our training setup in more detail.

\textbf{Model:} We use a 6-layer CNN with a VGG-16 like architecture. Specifically, the model is configured as [64 conv, 64 conv, Maxpool, 128 conv, 128 conv, Maxpool, 256 conv, 256 conv, Maxpool], where ``N conv'' indicates a convolutional layer with N number of 3$\times$3 filters and ``Maxpool'' indicates a Maxpool layer with kernel size 2$\times$2 and stride 2. Each convolutional layer is followed by a ReLU activation layer. 

Since quadratic regularizers focus on finding a model parameterization useful for next task while staying close to the previous parameterization, it is important the initial parameterization have high discriminative abilities. Thus, we follow prior works on quadratic regularization (\cite{MAS, SI}) and pretrain our model on CIFAR-10. 

\textbf{Classifier Heads:} Following the setup proposed by original works on quadratic regularization and also the other relevant baselines evaluated in our experiments, we use a multi-head setting for all experiments (\cite{EWC, MAS, SI, RWalk, agem, gem}).  

We do highlight that in a multi-head setting, with simple datasets (such as, MNIST) and only a few classes per task so that even random performance is quite high (e.g., only 2 classes per task or 50\% random performance), most continual/lifelong learning algorithms can perform well (as pointed out by \cite{cl_eval}). However, by using datasets of varying complexity, with both balanced/unbalanced classes, large/few number of samples per class, and enough classes per task so that random performance is sufficiently low (10--16\%), our experimental setup circumvents these issues. 

\textbf{Preprocessing:} Following general evaluation protocols (\cite{raghu}), we do not use data augmentation in our experiments. All samples are preprocessed to have a mean of [0.5, 0.5, 0.5] and standard deviation of [0.5, 0.5, 0.5] using the dataloader.

\textbf{Training/Evaluation Protocol:} All datasets are trained using SGD with momentum of 0.9. Other relevant hyperparameters are determined using a grid search (see below). We train for only 1 epoch per task, with a batch-size of 10, as is expected in streaming data settings \cite{gem}. The first 3 tasks are reserved for hyperparameter search and the remaining tasks are used to train the model. Only the final model is evaluated on the test splits for all datasets. 

\textbf{Hyperparameters:} We follow prior work (\cite{agem}) and use the first 3 tasks for all datasets to search for training hyperparameters when needed (e.g., for all experiments in Section 6). We use grid-search for finding the hyperparameters, with the following grids: 
\begin{itemize}
\item Learning rate ($\eta$): [0.1, 0.03, 0.01, 0.003, 0.001];
\item Regularization constant ($\lambda$): [$10^{-5}$, $10^{-4}$, $10^{-3}$, $10^{-2}$, $10^{-1}$, 1, $10^{1}$, $10^{2}$, $10^{3}$, $10^{4}$].
\end{itemize}

Note our grid for $\lambda$ is much wider than prior works (\cite{gem, agem}), which showed quadratic reguarlization approaches fail to achieve high performance. Our work demonstrates the high sensitivity of quadratic regularizers to hyperparameters. We thus note the use of a limited range of hyperparameters in prior works may also have been a factor in poor performance for quadratic regularizers.

The hyperparameters for different methods and datasets are reported in \autoref{tab:params} and \autoref{tab:params_30 epochs}. For A-GEM and ER-Reservoir, we use a memory buffer of 500 samples.

%% file: supplementary/files/sec4.tex
\section{More Results on Experiments from \autoref{sec:limitations}}
\label{sec:sec4}
\autoref{sec:limitations} of the main paper uses models trained on CIFAR-100 to demonstrate training instability issues caused by hyperparamter sensitivity and the impact of biased importance definitions on the effectiveness of popular quadratic regularizers. In this section, we repeat experiments from \autoref{sec:limitations} on Oxford-Flowers and Caltech-256.  Note that all results in the paper and in this section have been averaged over five seeds. 

\subsection{Case 1: $\eta \lambda \alpha_{n-1}^{(k)} < 0$.}
\label{sec41}
\begin{figure}[H]
\begin{center}
\begin{subfigure}{.44\textwidth}
	\centerline{\includegraphics[width=\columnwidth]{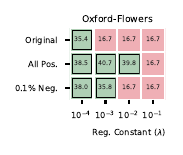}}
	\caption{Oxford-Flowers}
\end{subfigure}%
\begin{subfigure}{.44\textwidth}
	\centerline{\includegraphics[width=\columnwidth]{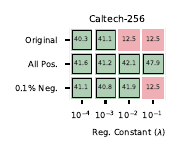}}
	\caption{Caltech-256}
\end{subfigure}
\caption{Impact of negative importance scores. Green, outlined cells indicate stable training; red cells indicate stable training. \emph{Original} implies both positive/negative scores are allowed, \emph{All Pos.} implies all scores are positive, and \emph{0.1\% Neg.} implies only 0.1\%, randomly picked parameters have negative scores. Average accuracy is mentioned in figure cells. Random accuracy is $10\%$ for CIFAR-100, $16.7\%$ for Oxford-Flowers, and $10\%$ for Caltech-256. Plain fine-tuning accuracy is $55.3\%$ for CIFAR-100, $37.9\%$ for Oxford-Flowers, and $40.1\%$ for Caltech-256. For all positive scores, training is always stable. Including even a few negative importance parameters often produces unstable training and training is rarely stable when both any number of negative importance parameters are allowed. Using a small $\lambda$ mitigates this behavior, but reduces the regularizer's effectiveness, resulting in similar performance to plain fine-tuning.}
\label{fig:negmaps_all}
\end{center}
\end{figure}

For parameters with negative importance, the quadratic regularizer enters the extrapolation regime and witnesses unstable training (see \textbf{Case~1} in \autoref{sec:lim1} of main paper). In \autoref{fig:negmaps_all}, we show this behavior holds true for all datasets considered in this work.

\subsection{Case 2: $\eta \lambda \alpha_{n-1}^{(k)} > 1$}
\label{sec42}
\begin{figure}[H]
\centering
\centerline{\includegraphics[width=\columnwidth]{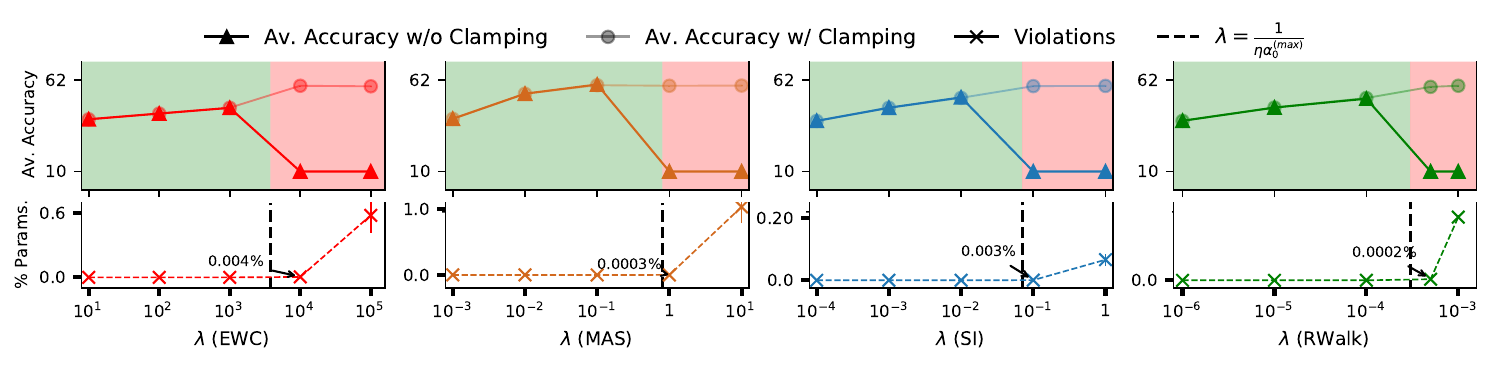}}
\caption{Oxford-Flowers: Number of violations of the inequality $\eta \lambda \alpha_{n-1}^{(k)} < 1$. The minimum number of violations at which we first see unstable training is noted as the lower limit on Y-axis of the plots (stable shaded green; unstable shaded red). As seen, across 1.2 million parameters, even a small number of violations result in unstable training: 35 (0.004\% parameters) for EWC, 4 (0.0003\% parameters) for MAS, 8 (0.0006\% parameters) for SI, and 17 (0.001\% parameters) for RWalk.}
\label{fig:violations_flowers}
\end{figure}

\begin{figure}[H]
\centering
\centerline{\includegraphics[width=\columnwidth]{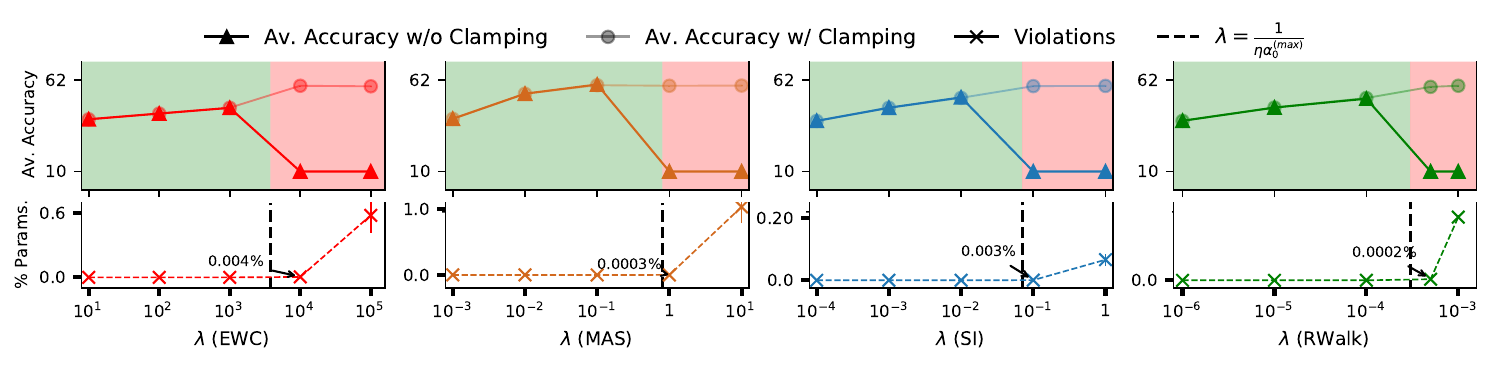}}
\caption{Caltech-256: Number of violations of the inequality $\eta \lambda \alpha_{n-1}^{(k)} < 1$. The minimum number of violations at which we first see unstable training is noted as the lower limit on Y-axis of the plots (stable shaded green; unstable shaded red). As seen, across 1.2 million parameters, even a small number of violations result in unstable training: 1 (0.0001\% parameters) for EWC, 4 (0.0003\% parameters) for MAS, 3 (0.0004\% parameters) for SI, and 6 (0.0005\% parameters) for RWalk.}
\label{fig:violations_cal256}
\end{figure}

When $\eta \lambda \alpha_{n-1}^{(k)} > 1$ for a parameter, the weights for interpolation becomes negative ($1 - \eta \lambda \alpha_{n-1}^{(k)} < 0$). This pushes the quadratic regularizer to the extrapolation regime, resulting in unstable training (see \textbf{Case~2} in \autoref{sec:lim1} of main paper). In \autoref{fig:violations} (CIFAR-100), \autoref{fig:violations_flowers} (Oxford-Flowers), and \autoref{fig:violations_cal256} (Caltech-256), we show this behavior holds true for all datasets considered in this work. 


\subsection{Disparate Importance Assignment}
\label{sec43}
\begin{figure}[H]
\begin{center}
\begin{subfigure}{.25\textwidth}
  \centering
  \centerline{\includegraphics[width=\columnwidth]{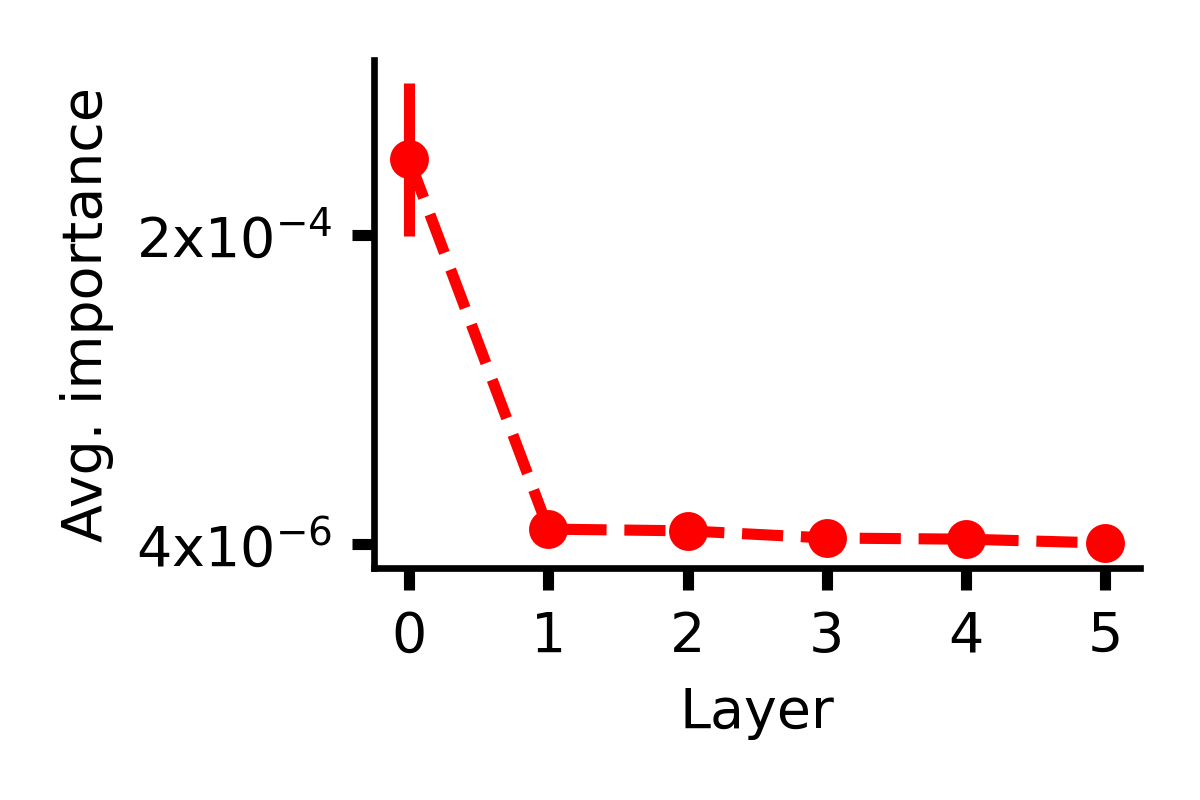}}
  \caption{EWC.}
\end{subfigure}%
\begin{subfigure}{.25\textwidth}
  \centering
  \centerline{\includegraphics[width=\columnwidth]{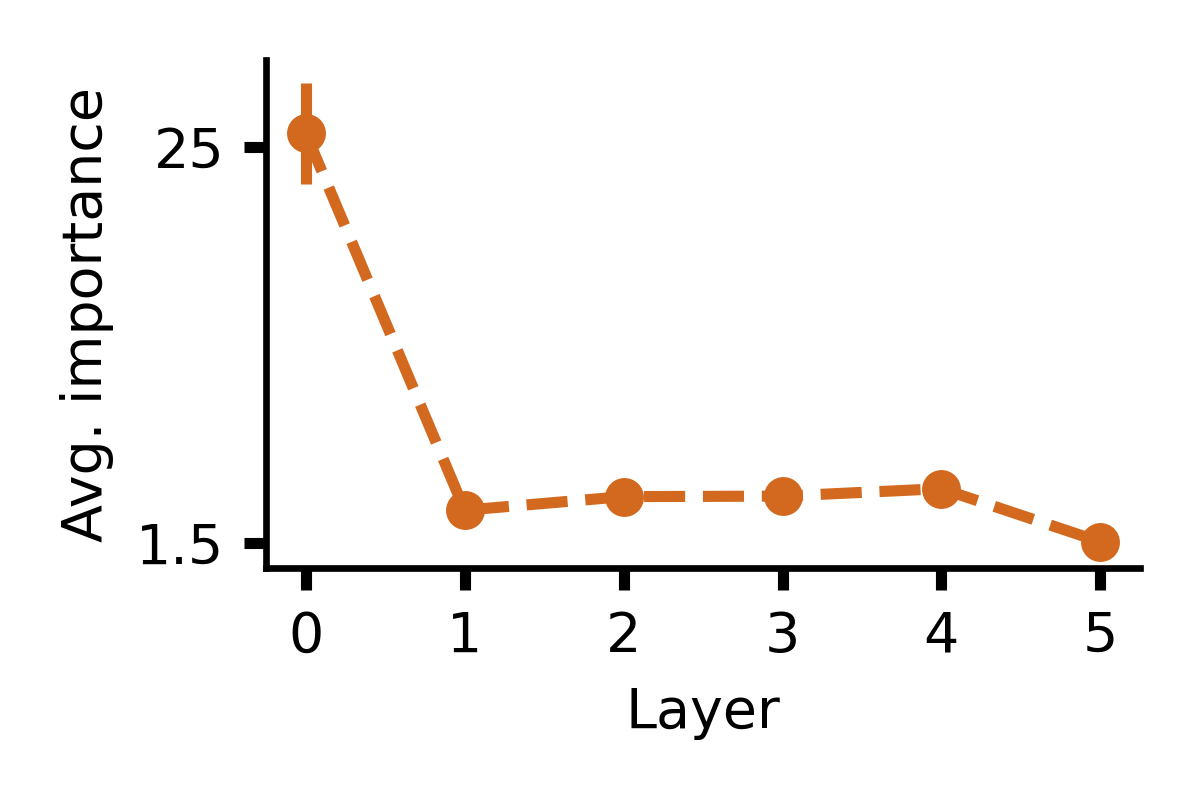}}
  \caption{MAS.}
\end{subfigure}%
\begin{subfigure}{.25\textwidth}
  \centering
  \centerline{\includegraphics[width=\columnwidth]{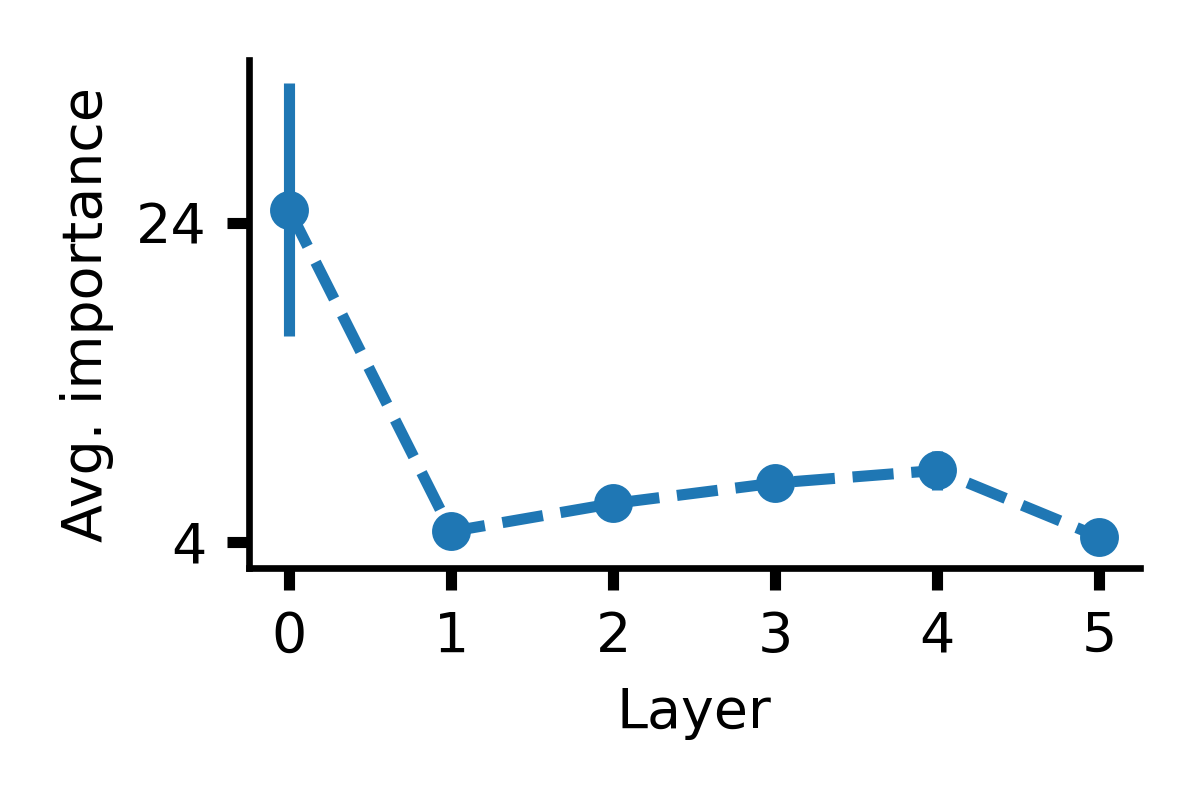}}
  \caption{SI.}
\end{subfigure}%
\begin{subfigure}{.25\textwidth}
  \centering
  \centerline{\includegraphics[width=\columnwidth]{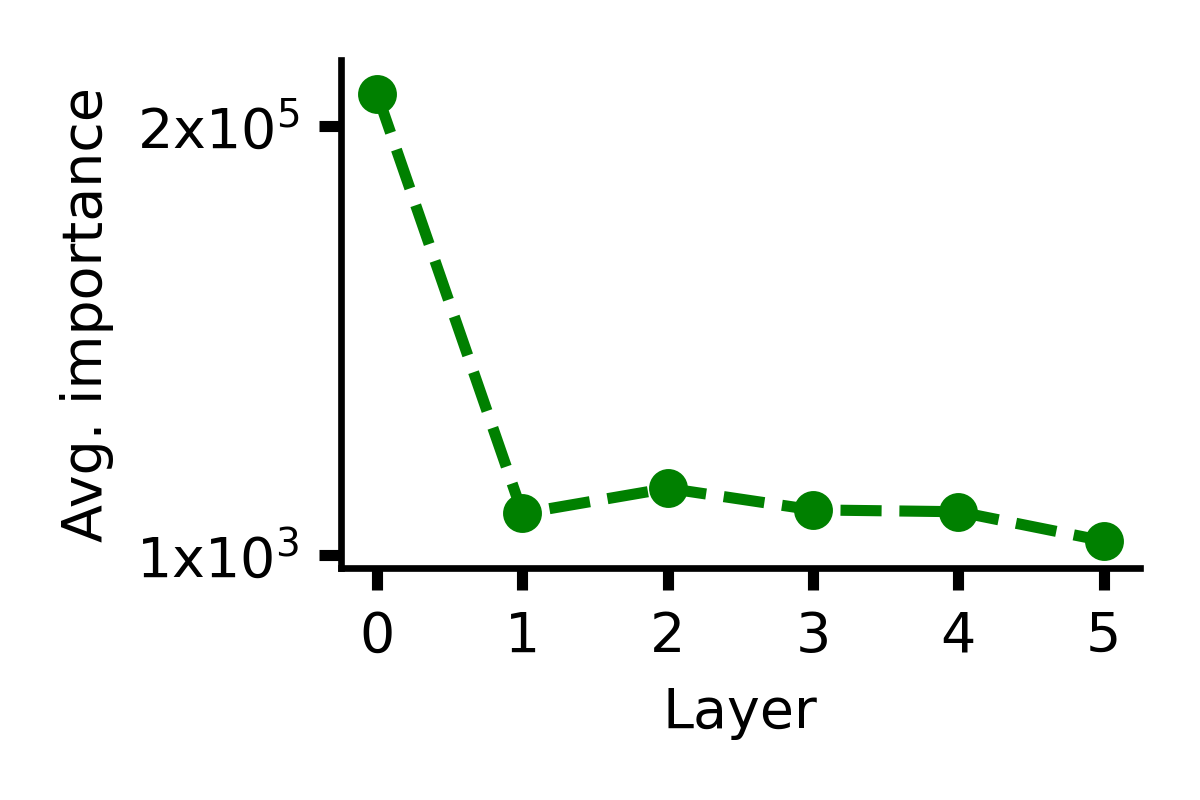}}
  \caption{RWalk.}
\end{subfigure}
\caption{Oxford-Flowers: Average importance of parameters in a layer.}
\label{fig:imp_scale_flowers}
\end{center}
\end{figure}

\begin{figure}[H]
\begin{center}
\begin{subfigure}{.25\textwidth}
  \centering
  \centerline{\includegraphics[width=\columnwidth]{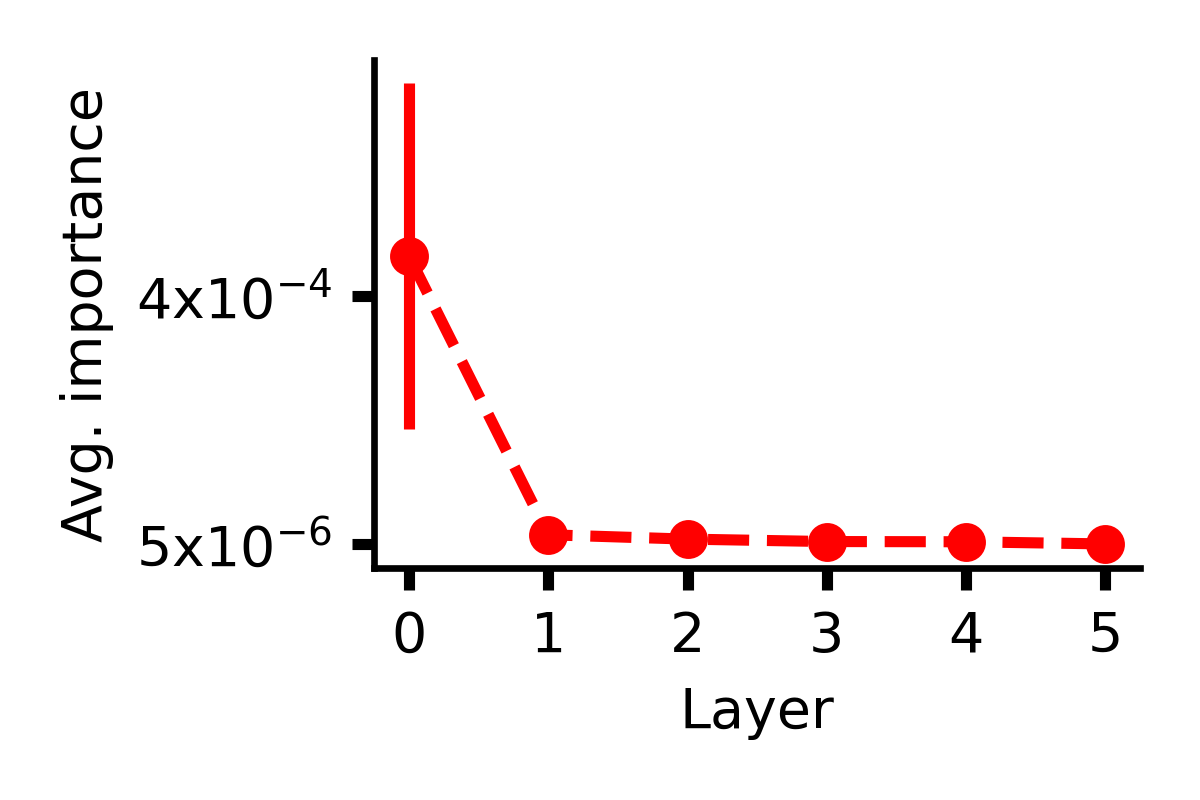}}
  \caption{EWC.}
\end{subfigure}%
\begin{subfigure}{.25\textwidth}
  \centering
  \centerline{\includegraphics[width=\columnwidth]{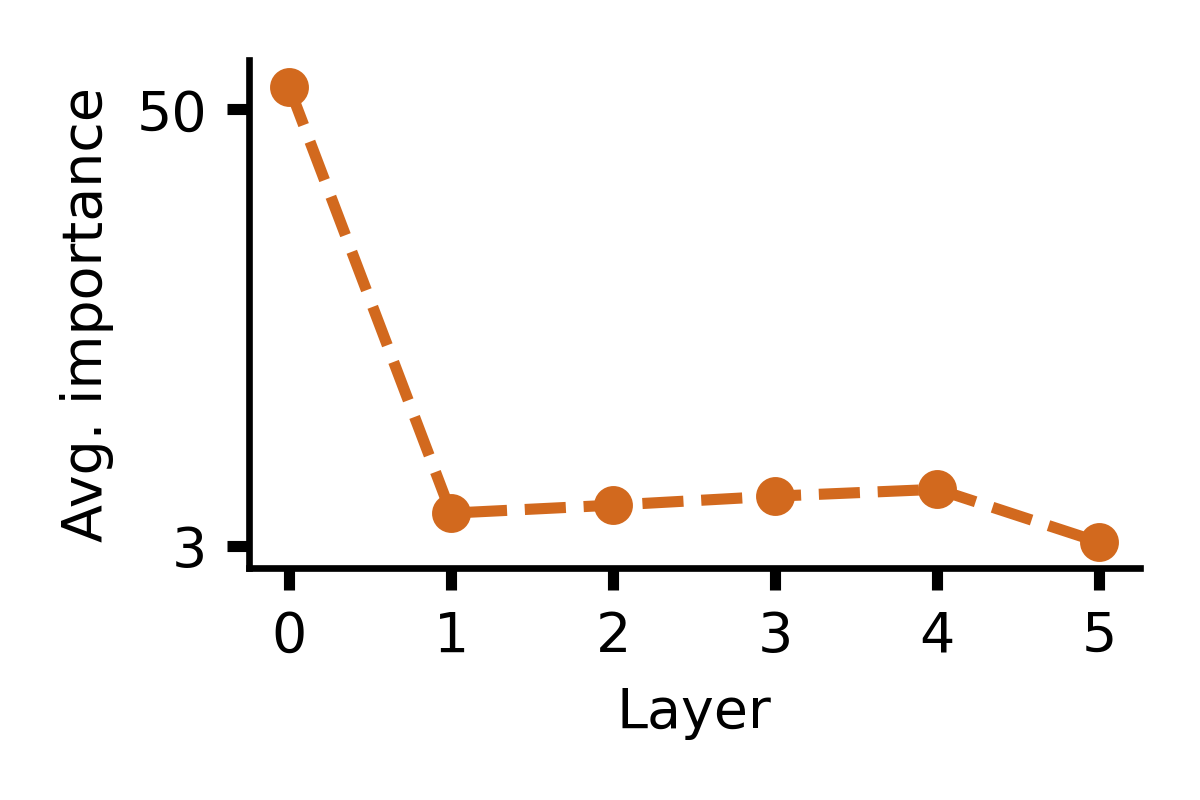}}
  \caption{MAS.}
\end{subfigure}%
\begin{subfigure}{.25\textwidth}
  \centering
  \centerline{\includegraphics[width=\columnwidth]{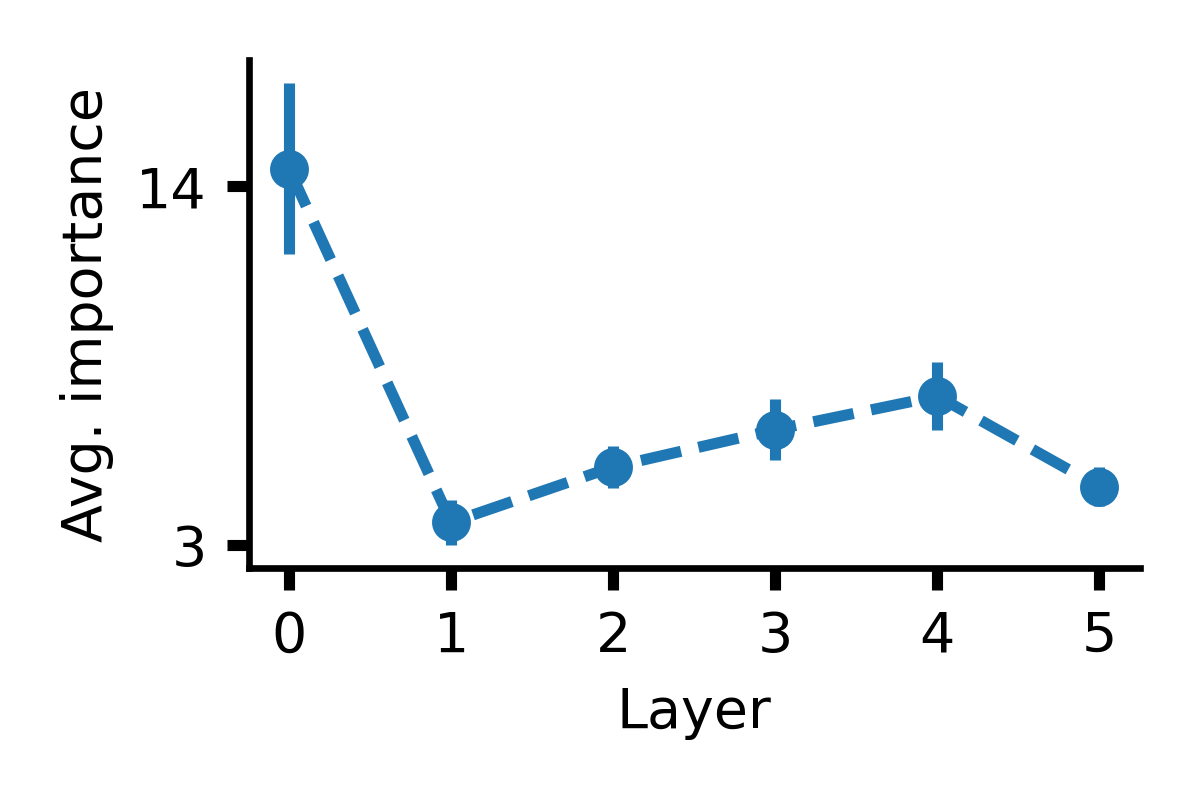}}
  \caption{SI.}
\end{subfigure}%
\begin{subfigure}{.25\textwidth}
  \centering
  \centerline{\includegraphics[width=\columnwidth]{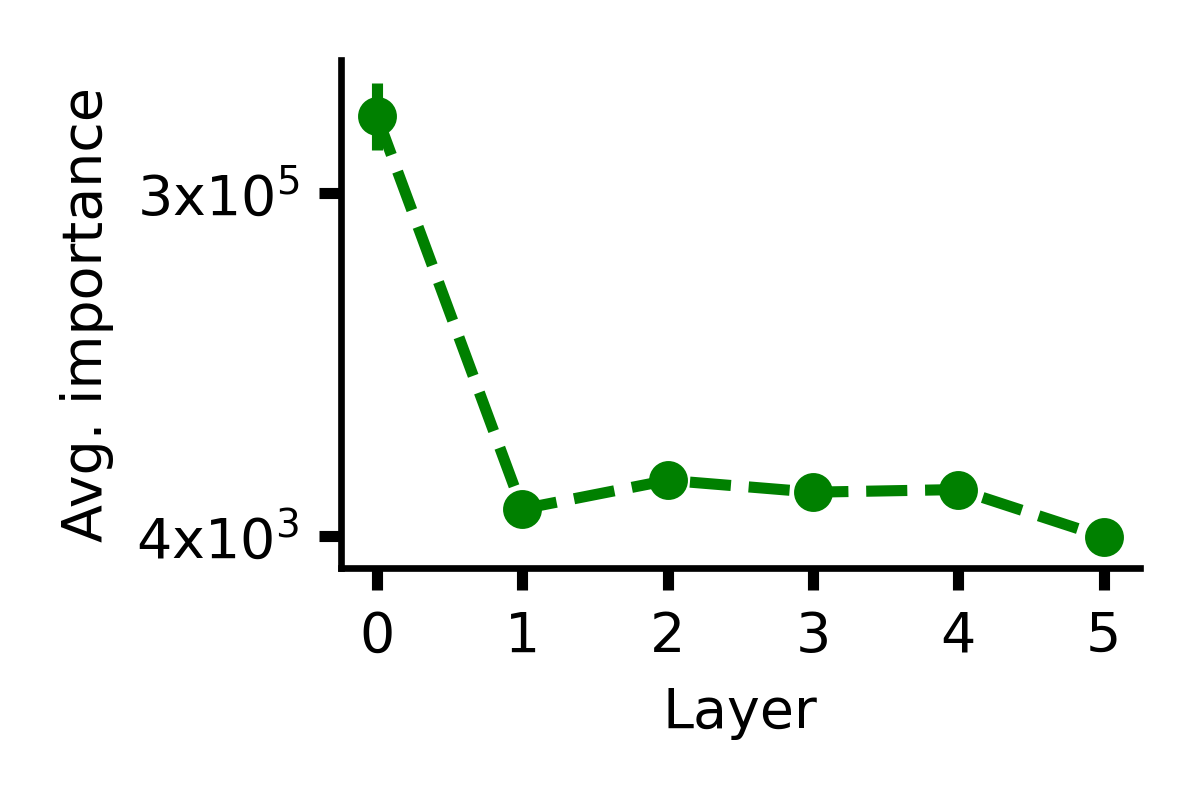}}
  \caption{RWalk.}
\end{subfigure}
\caption{Caltech-256: Average importance of parameters in a layer.}
\label{fig:imp_scale_cal256}
\end{center}
\end{figure}

Due to use of biased importance definitions, parameters in deeper layers are often assigned much lower importance than parameters in earlier layers \autoref{sec:lim2} of main paper). Thus, even if a valid training configuration (i.e., one that results in stable training) is used, the use of biased importance scores is unable to stop deeper layers from adapting to recent tasks. Since forgetting of previously learned tasks is majorly caused by changes in deeper layers, this disparate importance assignment renders quadratic regularizers ineffective. In \autoref{fig:imp_scale} (CIFAR-100), \autoref{fig:imp_scale_flowers} (Oxford-Flowers), and \autoref{fig:imp_scale_cal256} (Caltech-256), we show this disparate importance assignment is observed for all datasets considered in this work.


\begin{figure}[H]
\begin{center}
\begin{subfigure}{.49\textwidth}
	\centerline{\includegraphics[width=\columnwidth]{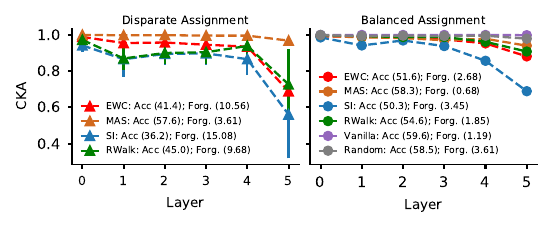}}
	\caption{Oxford-Flowers (1 epoch)}
\end{subfigure}%
\begin{subfigure}{.49\textwidth}
	\centerline{\includegraphics[width=\columnwidth]{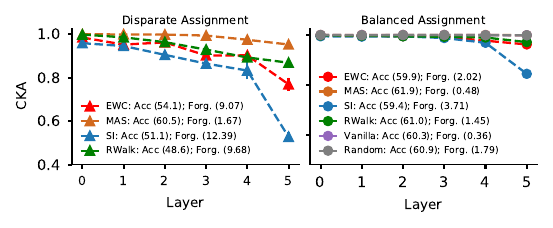}}
	\caption{Oxford-Flowers (30 epochs)}
\end{subfigure}
\begin{subfigure}{.49\textwidth}
	\centerline{\includegraphics[width=\columnwidth]{images/cka_analysis_flowers_1epoch.pdf}}
	\caption{Caltech-256 (1 epoch)}
\end{subfigure}%
\begin{subfigure}{.49\textwidth}
	\centerline{\includegraphics[width=\columnwidth]{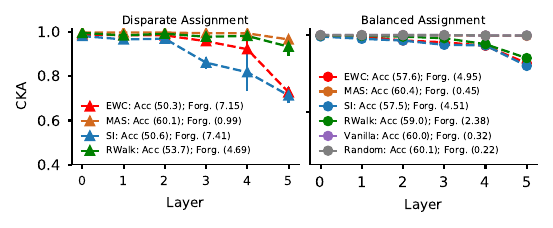}}
	\caption{Caltech-256 (30 epochs)}
\end{subfigure}
\caption{Representational similarity, as measured using CKA, between a model trained only on the first task of a dataset versus the model trained on all tasks of that dataset. Average accuracy (Acc.) and average forgetting (Forg.) are reported in the legend. Unlike other methods, Vanilla and Random are not biased against deeper layers, achieving high representational similarity across all layers with minimal forgetting. Meanwhile, the other measures suffer significant drift in similarity for deeper layers and, hence, more forgetting.}
\label{fig:cka_analysis_appendix}
\end{center}
\end{figure}

\underline{\smash{Balanced Importance Scores Prevent Forgetting More Effectively:}} In the main paper, we propose two regularizers that assign either unit importance score to all parameters (called Vanilla) or use uniformly picked, random importance score for all parameters (called Random). By treating all layers on a similar importance scale, we show these regularizers address problems pertaining to biased importance assignment in popular regularizes (see Figure~4 of main paper). In particular, following \cite{raghu}, we use CKA (\cite{CKA}) to measure representational similarity and estimate the contribution of a layer towards catastrophic forgetting. We find features across all layers of the model trained only on first task of a dataset and the model trained on all tasks of that dataset show high representational similarity for both Vanilla and Random. Here, we demonstrate this conclusion holds well for other datasets as well (see \autoref{fig:cka_analysis_appendix}).

\section{More Results on Experiments from \autoref{sec:exptt_eval}}
\label{sec:sec5}
\begin{table}
\scriptsize
\centering
\caption{Hyperparamaters found using grid search for different methods and datasets.}
\label{tab:res_params}
\begin{tabular}{@{}c|cccc|cccc@{}}
\toprule
Dataset & \multicolumn{4}{c|}{Plain Quadratic Regularizers}                        & \multicolumn{4}{c}{Explicit Interpolation Variants} \\ \midrule
CIFAR   & EWC   & MAS    & SI    & RWalk   & EWC   & MAS   & SI    & RWalk  \\ \midrule
LR ($\eta$)      & 0.003 & 0.003  & 0.003 & 0.003   & 0.01 & 0.01  & 0.01 & 0.01 \\
Reg ($\lambda$)     & $10^{4}$   & $10^{0}$    & $10^{-4}$ & $10^{-5}$    & -     & -     & -     & -         \\ \midrule
Flowers & EWC   & MAS    & SI    & RWalk   & EWC   & MAS   & SI    & RWalk  \\ \midrule
LR ($\eta$)      & 0.01 & 0.01  & 0.01 & 0.01   & 0.03 & 0.03 & 0.03 & 0.03 \\
Reg ($\lambda$)     & $10^{3}$   & $10^{0}$    & $10^{-2}$ & $10^{-5}$    & -     & -     & -     & -         \\ \midrule
Cal-256 & EWC   & MAS    & SI    & RWalk   & EWC   & MAS   & SI    & RWalk  \\ \midrule
LR ($\eta$)      & 0.01 & 0.01  & 0.01 & 0.01   & 0.03 & 0.03 & 0.03 & 0.03 \\
Reg ($\lambda$)     & $10^{3}$   & $10^{0}$    & $10^{-2}$ & $10^{-5}$    & -     & -     & -     & -         \\ \midrule
\end{tabular}
\end{table}

\subsection{Quadratic Regularization on ResNet-18}
\label{sec51}
In the main paper, we only provide results comparing quadratic regularization variants on a 6-layer CNN. Here, we give comparisons for a ResNet-18 model. Following prior works, we use one-third the number of filters in each layer (\cite{lmc}). Hyperparameters are again determined via a grid-search on first 3 tasks and listed in \autoref{tab:res_params}. The hyperparameter search space and other training configurations are the same as before (discussed in \autoref{sec:setup_appendix}). The final results are provided in \autoref{tab:res_results}. As can be seen, the conclusions drawn on 6-layer CNN hold well with the ResNet-18 model too. We note the results for plain fine-tuning can sometimes be better than EWC. This happens because grid-search leads to a small learning rate for plain fine-tuning, but a relatively larger learning rate for EWC. 

\subsection{Quadratic Regularization on 6-layer CNN}
\label{sec52}
In the main paper, we provided results comparing plain quadratic regularizers with their explicit interpolation variants. Due to space constraints, we did not provide standard deviations. This information, along with the average results, is shown in \autoref{tab:6_results}.

\begin{table}
\scriptsize
\centering
\caption{\label{tab:6_results} \textbf{Detailed version of Table 1 from main paper}: Comparison of Average Accuracy (Acc; $\uparrow$ indicates higher is better) and Average Forgetting (Forg; $\downarrow$ indicates lower is better) for plain and explicit interpolation variants EWC, MAS, SI, and RWalk. We consider three datasets: CIFAR-100 (10 tasks); Oxford-Flowers (17 tasks); and Caltech-256 (32 tasks). For a specific regularizer, the better performing variant is in bold. As shown, variants with explicit interpolation steps consistently outperform their Quadratic Regularization counterparts. This behavior is most prominent in datasets with unbalanced classes and longer task sequences (Oxford-Flowers and Caltech-256), where hyperparameter tuning is difficult.}
\begin{subtable}{\textwidth}
\centering
\begin{tabular}{@{}c!{\vrule width 1pt}c!{\vrule width 1pt}cccc!{\vrule width 1pt}cccc@{}}
\toprule
\multicolumn{2}{c!{\vrule width 1pt}}{1 epoch}       & \multicolumn{4}{c!{\vrule width 1pt}}{Plain Quadratic Regularizers}                                 & \multicolumn{4}{c}{Explicit Interpolation Variants} \\ \midrule
CIFAR   & Plain & EWC  & MAS  & SI   & RWalk      & EWC        & MAS                 & SI         & RWalk      \\ \midrule
Acc ($\uparrow$)    & 55.3 $\pm$ 1.6 & 62.5 $\pm$ 0.6 & 63.9 $\pm$ 0.8 & 61.2 $\pm$ 2.3 & {\bf 63.9 $\pm$ 0.3} & {\bf 63.8 $\pm$ 0.4} & {\bf 64.0 $\pm$ 0.2} & {\bf 63.9 $\pm$ 0.3} & 63.8 $\pm$ 0.9     \\
Forg ($\downarrow$)   & 9.13 $\pm$ 1.6 & 2.07 $\pm$ 1.0 & 0.56 $\pm$ 0.2 & 3.20 $\pm$ 0.3 & {\bf 0.54 $\pm$ 0.1}      & {\bf 0.08 $\pm$ 0.02} & {\bf 0.06 $\pm$ 0.05} & {\bf 0.07 $\pm$ 0.04} & {\bf 0.09 $\pm$ 0.03} \\ \midrule
Flowers & Plain & EWC  & MAS  & SI   & RWalk      & EWC        & MAS                 & SI         & RWalk      \\ \midrule
Acc ($\uparrow$)    & 37.9 $\pm$ 2.1 & 41.3 $\pm$ 0.5 & 57.5 $\pm$ 2.4 & 36.2 $\pm$ 2.3 & 44.9 $\pm$ 2.1      & {\bf 60.0 $\pm$ 0.8} & {\bf 59.6 $\pm$ 1.1}          & {\bf 59.9 $\pm$ 0.9} & {\bf 59.4 $\pm$ 1.1} \\
Forg ($\downarrow$)   & 13.4  $\pm$ 2.5  & 10.6  $\pm$ 1.0 & 3.61 $\pm$ 0.7 & 15.1 $\pm$ 1.9 & 9.68 $\pm$ 2.3      & {\bf 1.03 $\pm$ 0.5} & {\bf 0.44 $\pm$ 0.2}          & {\bf 0.99 $\pm$ 0.2} & {\bf 1.31 $\pm$ 0.3} \\ \midrule
Cal-256 & Plain & EWC  & MAS  & SI   & RWalk      & EWC        & MAS                 & SI         & RWalk      \\ \midrule
Acc ($\uparrow$)    & 40.1 $\pm$ 0.9  & 41.9 $\pm$ 0.5 & 53.7 $\pm$ 0.5 & 40.9 $\pm$ 0.5 & 43.7 $\pm$ 0.7      & {\bf 56.2 $\pm$ 0.3} & {\bf 57.3 $\pm$ 0.6} & {\bf 56.0  $\pm$ 0.6} & {\bf 55.8  $\pm$ 0.5} \\
Forg ($\downarrow$)   & 6.21 $\pm$ 0.5 & 5.59 $\pm$ 0.1 & 3.92 $\pm$ 0.4 & 5.68 $\pm$ 0.1 & 4.69 $\pm$ 0.1      & {\bf 1.41 $\pm$ 0.2} & {\bf 0.36  $\pm$ 0.1} & {\bf 1.42  $\pm$ 0.2} & {\bf 1.31 $\pm$ 0.1} \\ \bottomrule
\end{tabular}
\end{subtable}
\begin{subtable}{\textwidth}
\centering
\begin{tabular}{@{}c!{\vrule width 1pt}c!{\vrule width 1pt}cccc!{\vrule width 1pt}cccc@{}}
\toprule
\multicolumn{2}{c!{\vrule width 1pt}}{30 epochs}       & \multicolumn{4}{c!{\vrule width 1pt}}{Plain Quadratic Regularizers}                                 & \multicolumn{4}{c}{Explicit Interpolation Variants} \\ \midrule
CIFAR   				& Plain 			& EWC  				& MAS  				& SI   				& RWalk      				& EWC        			& MAS                 	& SI         			& RWalk      \\ \midrule
Acc ($\uparrow$)    	& 60.7 $\pm$ 1.4 	& 63.6 $\pm$ 0.8 	& 66.3 $\pm$ 0.2 	& 62.8 $\pm$ 1.6 	& 64.7 $\pm$ 0.9 			& {\bf 66.3 $\pm$ 0.3} 	& {\bf 66.0 $\pm$ 0.2} 	& {\bf 66.2 $\pm$ 0.3} 	& {\bf 66.1 $\pm$ 0.3}     \\
Forg ($\downarrow$)   	& 8.01 $\pm$ 1.2 	& 4.63 $\pm$ 1.0 	& 0.23 $\pm$ 0.07 	& 6.30 $\pm$ 0.7 	& 3.39 $\pm$ 1.1 		    & {\bf 0.21 $\pm$ 0.06} & {\bf 0.23 $\pm$ 0.07} & {\bf 0.25 $\pm$ 0.04} & {\bf 0.29 $\pm$ 0.07} \\ \midrule
Flowers 				& Plain 			& EWC  				& MAS  				& SI   				& RWalk      				& EWC        			& MAS                 	& SI         			& RWalk      \\ \midrule
Acc ($\uparrow$)    	& 49.9 $\pm$ 1.7 	& 54.1 $\pm$ 0.9 	& 60.5 $\pm$ 0.6 	& 51.1 $\pm$ 1.4 	& 55.7 $\pm$ 1.5      		& {\bf 61.3 $\pm$ 0.7} 	& {\bf 61.6 $\pm$ 0.6}  & {\bf 61.8 $\pm$ 0.7} 	& {\bf 61.7 $\pm$ 0.8} \\
Forg ($\downarrow$)   	& 11.8  $\pm$ 1.9  	& 9.07  $\pm$ 0.8 	& 1.19 $\pm$ 0.7 	& 12.4 $\pm$ 2.0 	& 7.19 $\pm$ 1.3      		& {\bf 0.21 $\pm$ 0.4} 	& {\bf 0.48 $\pm$ 0.5}  & {\bf 0.62 $\pm$ 0.7} 	& {\bf 0.71 $\pm$ 0.5} \\ \midrule
Cal-256 				& Plain 			& EWC  				& MAS  				& SI   				& RWalk      				& EWC        			& MAS                 	& SI         			& RWalk      \\ \midrule
Acc ($\uparrow$)    	& 40.6 $\pm$ 1.2  	& 50.3 $\pm$ 0.9 	& 60.1 $\pm$ 0.4 	& 50.6 $\pm$ 0.8 	& 53.7 $\pm$ 0.8      		& {\bf 60.6 $\pm$ 1.0} 	& {\bf 60.9 $\pm$ 0.8} 	& {\bf 60.8  $\pm$ 0.8} & {\bf 60.6  $\pm$ 0.7} \\
Forg ($\downarrow$)   	& 15.8 $\pm$ 1.5 	& 7.15 $\pm$ 0.8 	& 0.99 $\pm$ 0.4 	& 7.41 $\pm$ 1.1 	& 4.69 $\pm$ 0.6      		& {\bf 0.22 $\pm$ 0.6} 	& {\bf 0.18  $\pm$ 0.4} & {\bf 0.32  $\pm$ 0.6} & {\bf 0.34 $\pm$ 0.6} \\ \bottomrule
\end{tabular}
\end{subtable}
\end{table}

\begin{table}
\scriptsize
\centering
\caption{\label{tab:res_results} \textbf{Results on ResNet-18}: Comparison of Average Accuracy (Acc; $\uparrow$ indicates higher is better) and Average Forgetting (Forg; $\downarrow$ indicates lower is better) for plain and explicit interpolation variants EWC, MAS, SI, and RWalk. We consider three datasets: CIFAR-100 (10 tasks); Oxford-Flowers (17 tasks); and Caltech-256 (32 tasks). For a specific regularizer, the better performing variant is in bold. As shown, variants with explicit interpolation steps consistently outperform their Quadratic Regularization counterparts. This behavior is most prominent in datasets with unbalanced classes and longer task sequences (Oxford-Flowers and Caltech-256), where hyperparameter tuning is difficult.}
\begin{tabular}{@{}c!{\vrule width 1pt}c!{\vrule width 1pt}cccc!{\vrule width 1pt}cccc@{}}
\toprule
Dataset &       & \multicolumn{4}{c!{\vrule width 1pt}}{Plain Quadratic Regularizers}                                 & \multicolumn{4}{c}{Explicit Interpolation Variants} \\ \midrule
CIFAR   & Plain & EWC  & MAS  & SI   & RWalk      & EWC        & MAS                 & SI         & RWalk      \\ \midrule
Acc ($\uparrow$)    & 40.1 $\pm$ 0.6  & 39.6 $\pm$ 1.4 & \bf{58.5 $\pm$ 0.5} & 30.5 $\pm$ 1.8 & {\bf 58.6 $\pm$ 0.3} & {\bf 57.6 $\pm$ 0.4} & 57.4 $\pm$ 0.3 & {\bf 57.2 $\pm$ 0.6} & 57.3 $\pm$ 0.5       \\
Forg ($\downarrow$)   & 19.3 $\pm$ 0.2  & 33.2 $\pm$ 4.9 & \bf{1.7 $\pm$ 0.4} & 39.4 $\pm$ 1.3 & 4.2 $\pm$ 0.2       & {\bf 1.7 $\pm$ 0.3} & 1.7 $\pm$ 0.6 & {\bf 1.9 $\pm$ 0.3} & {\bf 1.6 $\pm$ 0.2} \\ \midrule
Flowers & Plain & EWC  & MAS  & SI   & RWalk      & EWC        & MAS                 & SI         & RWalk      \\ \midrule
Acc ($\uparrow$)    & 32.1 $\pm$ 1.7  & 27.3 $\pm$ 3.4 & 41.4 $\pm$ 1.5 & 24.6 $\pm$ 1.6 & 41.6 $\pm$ 0.6       & {\bf 53.4 $\pm$ 1.4} & {\bf 52.6 $\pm$ 1.8} & {\bf 52.5 $\pm$ 1.8} & {\bf 53.1 $\pm$ 1.7} \\
Forg ($\downarrow$)   & 9.6 $\pm$ 1.3  & 32.1 $\pm$ 3.4 & 16.9 $\pm$ 1.3 & 34.7 $\pm$ 1.5 & 15.4 $\pm$ 1.0   & {\bf 7.1 $\pm$ 1.2} & {\bf 7.9 $\pm$ 1.2}      & {\bf 5.1 $\pm$ 0.9} &  {\bf 8.4 $\pm$ 1.8} \\ \midrule
Cal-256 & Plain & EWC  & MAS  & SI   & RWalk      & EWC        & MAS                 & SI         & RWalk      \\ \midrule
Acc ($\uparrow$)    & 35.7 $\pm$ 1.1 & 24.6 $\pm$ 0.9 & 46.4  $\pm$ 1.4 & 22.9  $\pm$ 0.5 & 56.8  $\pm$ 2.6       & {\bf 52.7 $\pm$ 1.5} & {\bf 52.9 $\pm$ 1.3} & {\bf 50.0 $\pm$ 1.4} & {\bf 52.8 $\pm$ 1.7} \\
Forg ($\downarrow$)   & 13.7 $\pm$ 1.5  & 32.7 $\pm$ 1.2 & 12.9  $\pm$ 1.6 & 31.7 $\pm$ 1.0 & 12.1 $\pm$ 2.3       & {\bf 3.47 $\pm$ 0.3} & {\bf 3.8 $\pm$ 0.3} & {\bf 3.2 $\pm$ 0.7} & {\bf 3.8 $\pm$ 0.7} \\ \bottomrule
\end{tabular}
\end{table}